\documentclass[10pt,twocolumn,letterpaper]{article}

\usepackage[pagenumbers]{cvpr} 

\usepackage{graphicx}
\usepackage[table]{xcolor}
\usepackage{booktabs}
\usepackage{multirow}
\usepackage{arydshln}
\usepackage[accsupp]{axessibility}  

\definecolor{cvprblue}{rgb}{0.21,0.49,0.74}
\usepackage[pagebackref,breaklinks,colorlinks,allcolors=cvprblue]{hyperref}

\definecolor{darkgreen}{rgb}{0, 0.5, 0}
\definecolor{myblue}{RGB}{47, 114, 193}
\definecolor{brickred}{rgb}{0.8, 0.25, 0.33}
\definecolor{brandeisblue}{rgb}{0.0, 0.44, 1.0}
\definecolor{blueish}{rgb}{0.0, 0.3, 0.6}

\definecolor{orange}{HTML}{cc7700}
\definecolor{green}{HTML}{339955}
\definecolor{Highlight}{rgb}{0.12,0.49,0.85}
\definecolor{my_red}{HTML}{FE4444}
\definecolor{tab_green}{RGB}{224, 254, 220}
\definecolor{tab_blue}{RGB}{217, 217, 252}

\newcommand{\NAME}{SparVAR\xspace}

\def\paperID{*****} 
\def\confName{CVPR}
\def\confYear{2026}

\title{SparVAR: Exploring Sparsity in Visual AutoRegressive Modeling for Training-Free Acceleration}

\vspace{-4mm}
\author{%
\textbf{Zekun Li}$^{1,2,3*}$ \quad
\textbf{Ning Wang}$^{1,2,3*}$ \quad
\textbf{Tongxin Bai}$^{3\dagger}$ \quad
\textbf{Changwang Mei}$^{1,4}$ \\
\textbf{Peisong Wang}$^{1,2\dagger}$ \quad
\textbf{Shuang Qiu}$^{5}$ \quad
\textbf{Jian Cheng}$^{1,2\dagger}$ \\
    $^1$Institute of Automation, Chinese Academy of Sciences, \\
    $^2$School of Artificial Intelligence, University of Chinese Academy of Sciences, \\
    $^3$Beijing Academy of Artificial Intelligence, \\
    $^4$Nanjing University of Science and Technology, \\
    $^5$City University of Hong Kong. \\
}

\begin{document}

\renewcommand{\thefootnote}{\fnsymbol{footnote}}

\twocolumn[{%
\renewcommand\twocolumn[1][]{#1}%
\maketitle
\vspace{-13mm}
\begin{center}
    \captionsetup{type=figure}
    \includegraphics[width=\linewidth]{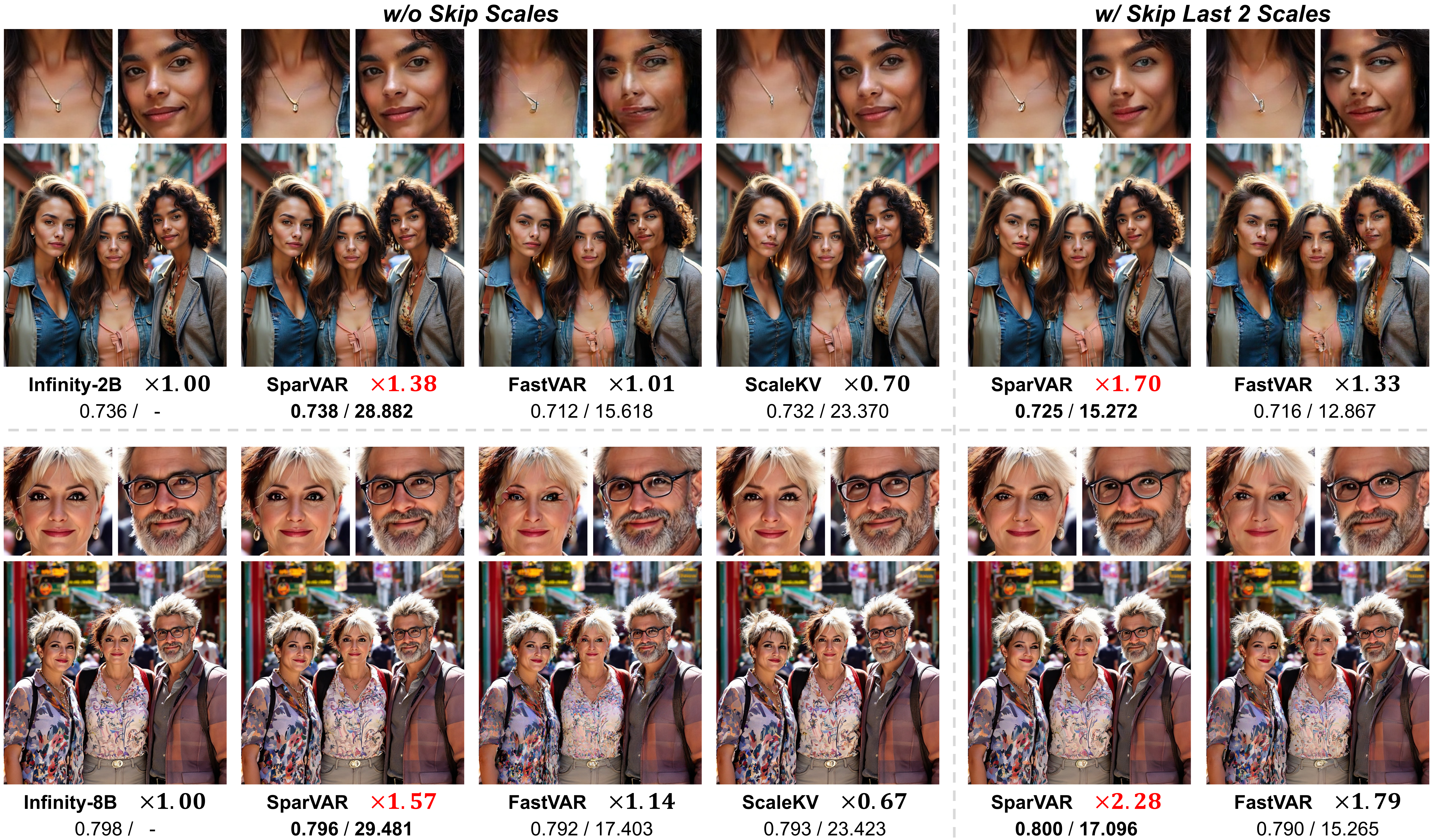}
    \vspace{-6mm}
    \captionof{figure}{
        Our \textbf{\NAME} achieves efficient acceleration while preserving high-frequency details consistent with the baseline~\cite{infinity}, whereas prior methods introduce visible artifacts and texture loss.
        %
        The bottom metrics denotes GenEval / PSNR.
        Zoom in for fine-detail comparison.
    }
    \vspace{-1mm}
    \label{fig:fig1}
\end{center}%
}]

\footnotetext[1]{Equal contribution.\quad\textsuperscript{\(\dagger\)}Corresponding authors.}


\begin{abstract}
Visual AutoRegressive (VAR) modeling has garnered significant attention for its innovative next-scale prediction paradigm.
However, mainstream VAR paradigms attend to all tokens across historical scales at each autoregressive step. 
As the next scale resolution grows, the computational complexity of attention increases quartically with resolution, causing substantial latency.
Prior accelerations often skip high-resolution scales, which speeds up inference but discards high-frequency details and harms image quality.
To address these problems, we present \textbf{\NAME}, a training-free acceleration framework that exploits three properties of VAR attention: \textbf{(i) strong attention sinks}, \textbf{(ii) cross-scale activation similarity}, and \textbf{(iii) pronounced locality}.
Specifically, we dynamically predict the sparse attention pattern of later high-resolution scales from a sparse decision scale, and construct scale self-similar sparse attention via an efficient index-mapping mechanism, enabling high-efficiency sparse attention computation at large scales.
Furthermore, we propose cross-scale local sparse attention and implement an efficient block-wise sparse kernel, which achieves $\mathbf{> 5\times}$ faster forward speed than FlashAttention.
Extensive experiments demonstrate that the proposed \NAME can reduce the generation time of an 8B model producing $1024\times1024$ high-resolution images to the \textbf{1s}, \textbf{without skipping the last scales}.
Compared with the VAR baseline accelerated by FlashAttention, our method achieves a $\mathbf{1.57\times}$ speed-up while preserving almost all high-frequency details. 
When combined with existing scale-skipping strategies, \NAME attains up to a $\mathbf{2.28\times}$ acceleration, while maintaining competitive visual generation quality.
Code is available at \href{https://github.com/CAS-CLab/SparVAR}{SparVAR}.
\end{abstract}

\begin{figure*}[!t]
    \centering
    \includegraphics[width=0.96\linewidth]{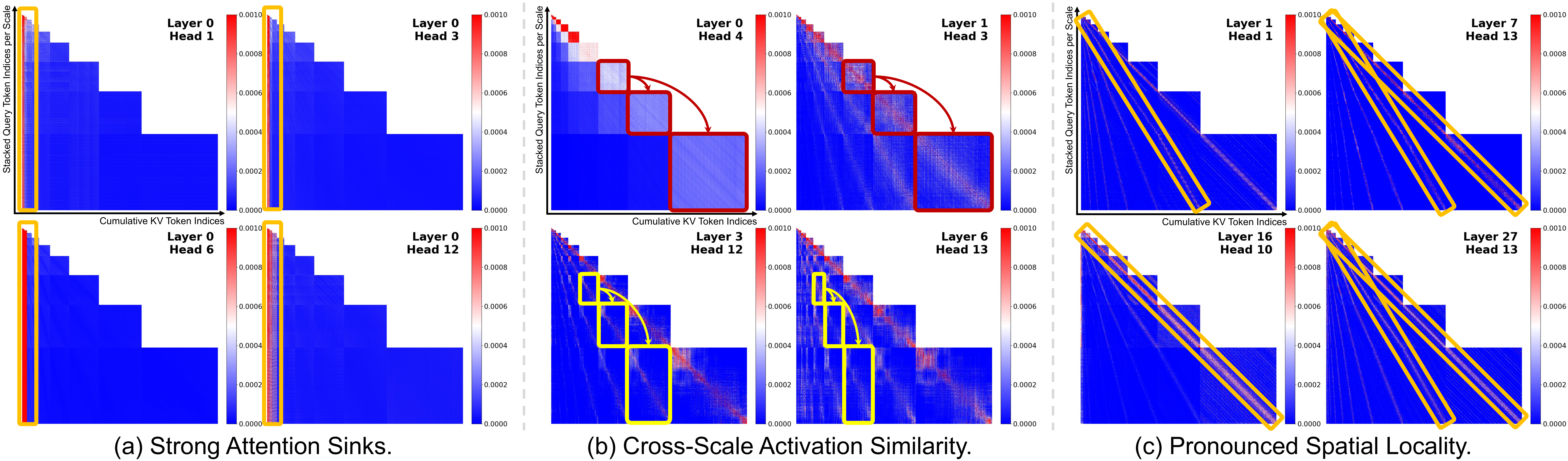}
    \vspace{-2mm}
    \caption{
        Visualization of attention activation patterns in the Infinity~\cite{infinity} across different layers and heads.
        \textbf{(a) Strong Attention Sinks}: early-scale tokens consistently attract large attention weights, serving as global anchors that dominate image structure formation.
        \textbf{(b) Cross-Scale Activation Similarity}: corresponding sub-blocks across adjacent scales exhibit similar activation distributions, indicating sparse attention patterns that can be transferred across scales.
        \textbf{(c) Pronounced Spatial Locality}: at larger scales, attention becomes increasingly concentrated along local spatial bands, revealing strong locality both within and between neighboring scales.
    }
    \label{fig:attn_map_3_feats}
    \vspace{-4mm}
\end{figure*}
\section{Introduction}
\label{sec:intro}

Recent advances in visual generative modeling have been dominated by two major paradigms: diffusion models~\cite{ho2020denoising,ldm,sdxl,dit,pixart,pixart_sigma,hunyuanvideo,open_sora} and autoregressive (AR) architectures~\cite{vqgan,llamagen,lumina-mgpt,mar,show-o,emu3}. 
Diffusion models achieve remarkable image quality via iterative denoising but suffer from slow generation due to long sampling chains, whereas AR models generate visual tokens sequentially~\cite{ar_vision_survey}, typically offering higher sampling efficiency~\cite{randar,par}. 
Among them, visual autoregressive (VAR) modeling~\cite{var,hart,infinity} has recently emerged as a promising alternative by introducing a novel next-scale prediction paradigm: instead of generating individual token at each step, VAR predicts all tokens of the next scale in parallel, progressively refining higher-resolution residuals~\cite{resnet} and enabling more efficient and scalable inference.

Despite its strong potential, current VAR frameworks still suffer from high inference latency when generating high-resolution images.
In next-scale prediction, tokens at the current scale must attend to all tokens from previous scales to maintain structural consistency, causing the attention complexity~\cite{attention} to grow roughly quartically ($\mathcal{O}(n^2) \to \mathcal{O}(n^4)$) with image resolution and making the last two large-scale steps account for about 60\% of the total runtime~\cite{fastvar}.
Meanwhile, the accumulation of key–value (KV) caches across historical scales drastically increases GPU memory usage---an 8B VAR model~\cite{infinity} requires nearly 60 GB to generate $1024\times1024$ images~\cite{scalekv}---thereby limiting batch inference, throughput, and deployment scalability.

To alleviate this bottleneck, recent acceleration techniques for VAR focus primarily on \textbf{skipping late high-resolution scales}.
By omitting the last two or three scales, methods such as FastVAR~\cite{fastvar} and SkipVAR~\cite{skipvar} substantially reduce computation and inference latency while maintaining semantic correctness.
However, this improvement comes \textbf{at the cost of losing fine-grained high-frequency details}, as the last scales are responsible for texture refinement, spatial sharpness and even detailed content generation.
As shown in Fig.~\ref{fig:fig1}, although high-level metrics such as GenEval~\cite{geneval} score may still indicate semantic alignment, low-level metrics---including Peak Signal-to-Noise Ratio (PSNR), Structural Similarity Index Measure (SSIM)~\cite{ssim}, and Learned Perceptual Image Patch Similarity (LPIPS)~\cite{lpips}---clearly reveal a substantial drop in visual fidelity compared with the baseline model Infinity~\cite{infinity}.
Even adaptive strategies like SkipVAR, which determine skipping dynamically per-sample, still yield noticeably lower low-level scores.
We further observe that in multi-object or fine-texture scenarios, skipping high-resolution scales often leads to \textbf{missing texture patterns} and \textbf{structural distortions} in the generated images.

These limitations raise a question:
\textbf{\textit{Can we achieve effective acceleration for VAR without skipping any scales?}}

\begin{figure}[t]
    \centering
    \includegraphics[width=1\linewidth]{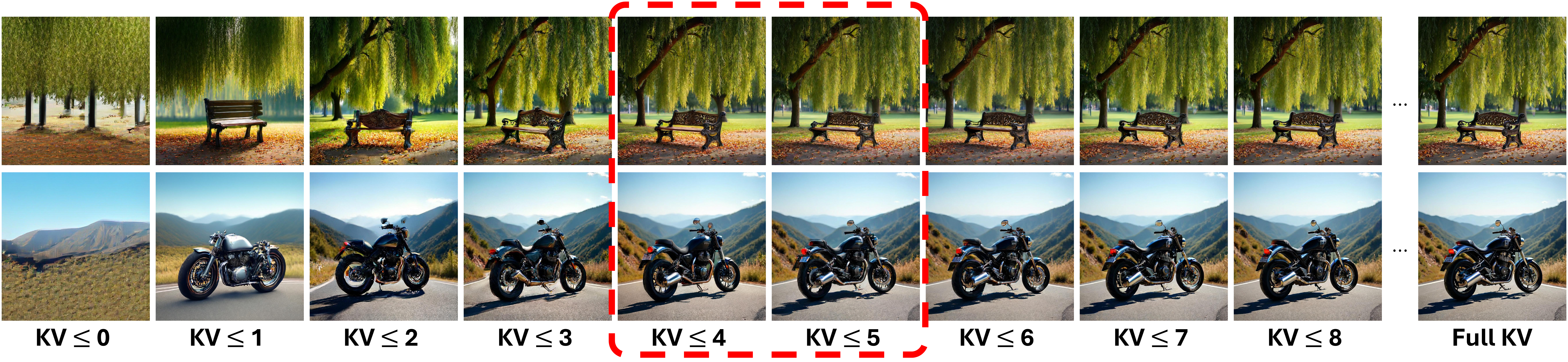}
    \vspace{-6mm}
    \caption{
        The influence of attention sinks in KV cache on generated images.
        Refer to the appendix for the large image.
    }
   \label{fig:kv_sink}
   \vspace{-6mm}
\end{figure}

To answer this question, we conduct a systematic analysis of the attention activation patterns in Infinity, a representative text-to-image VAR model, and reveal three consistent properties across layers, heads, and scales:
\textbf{\textit{1) Strong Attention Sinks.}}
As shown in Fig.~\ref{fig:attn_map_3_feats}(a), we observe that a small subset of early-scale tokens consistently attracts high attention weights, functioning as ``global anchors''~\cite{streamingllm}.
We further investigate the influence of historical scale KV caches on the final generation results: as illustrated in Fig.~\ref{fig:kv_sink}, when retaining only the first $4\sim 5$ scales of the KV cache, the model is still able to reconstruct accurate object layouts and global structures.
\textbf{\textit{2) Cross-Scale Activation Similarity.}}
As shown in Fig.~\ref{fig:attn_map_3_feats}(b), for instance, the attention of tokens within scale 10 (highlighted by the red box) closely resembles that of scales 11 and 12.
Moreover, in the cross-scale regions where tokens from the current scale interact with those from historical scales (yellow box), similar attention structures can also be observed.
This similarity indicates that the attention of the last scales can be effectively predicted from earlier scales, providing a solid foundation for efficient sparse computation at high-resolutions scales.
\textbf{\textit{3) Pronounced Spatial Locality.}}
As the resolution of next-scale increases, the attention patterns become increasingly concentrated within local spatial neighborhoods of the same or adjacent scales. 
In particular, this locality manifests as cross-scale diagonal activation patterns in the attention maps.
%
These systematic findings reveal substantial redundancy in the attention activation patterns of VAR, both intra- and cross-scale, inspiring us to develop a training-free sparse acceleration framework for efficient inference.

Building upon these observations, we introduce \textbf{\NAME}, a \textbf{training-free acceleration framework} that exploits the sparsity of VAR attention.
Specifically, we introduce two plug-and-play efficient attention modules.
1) \textbf{Cross-Scale Self-Similar Sparse Attention} ($CS^4A$), which dynamically predicts the sparse attention patterns of high-resolution scales from a sparse decision scale via an efficient cross-scale sparse index mapping.
\NAME transfers sparsity across scales and performs attention computation only over the selected sparse KV cache at large scales, significantly improving efficiency.
2) \textbf{Cross-Scale Local Sparse Attention} ($CSLA$), which reinforces spatial locality via an optimized \textbf{Block-wise Local Sparse Kernel} ($\mathbf{> 5\times}$ faster than Flash-attention~\cite{flashattention,flashattention2} in forward), achieving substantial computational savings while preserving fine-grained details.
Together, these components enable \NAME to accelerate inference without skipping any scales, allowing an 8B model to generate $1024\times1024$ images at second-level latency with a $\mathbf{1.57\times}$ speed-up. 
Most importantly, \NAME maintains semantic consistency and achieves nearly identical high-fidelity image quality compared with the baseline model.

Our main contributions are summarized as follows:
\begin{itemize}
    \item We systematically analyze the attention patterns in pretrained VAR models and observe three key phenomena---strong attention sinks, cross-scale activation similarity, and pronounced spatial locality---which directly motivate the exploration of attention sparsity in VAR.
    \item Based on these findings, we design two plug-and-play sparse attention modules and implement customized block-wise sparse kernels that achieve significant acceleration while preserving fine-grained generation details.
    \item The proposed \NAME is a training-free inference acceleration framework that exploits cross-scale attention sparsity in VAR, achieving up to $\mathbf{1.57\times}$ faster inference without sacrificing high-frequency fidelity.
\end{itemize}

\section{Related Work}
\label{sec:re_works}

\noindent
\textbf{Autoregressive Visual Generation.}
Early autoregressive (AR) image generators flatten a 2D image into a 1D raster-scan sequence and predict the next pixel~\cite{pixelcnn,gated_pixelcnn,pixelcnn++}, which becomes prohibitively expensive at high resolutions.
To improve scalability, subsequent work adopts vector quantization~\cite{vqvae,vqvae2,zhou2025tvc} to convert image patches into discrete tokens and pairs them with Transformers~\cite{attention} in a next-token paradigm~\cite{vqgan,llamagen,lumina-mgpt}.
Leveraging large language model scaling laws~\cite{gpt2,henighan2020scaling,kaplan2020scaling}, token-based AR achieves image quality competitive with state-of-the-art diffusion models~\cite{ldm,sdxl,pixart,pixart_sigma,esser2024scaling,flux_kontext,xue2025contemporary}.
However, strictly sequential token generation still incurs substantial latency and raster flattening weakens 2D inductive bias, making efficient high-resolution synthesis challenging.
Recently, visual autoregressive (VAR) modeling~\cite{var,hart,infinity} adopts a next-scale paradigm: at each AR step the model predicts all tokens of the next resolution scale in parallel, generating images in a coarse-to-fine manner.
Yet at each step VAR attention must attend to all tokens from previous scales, so the computational and memory costs grow rapidly with resolution.
This scalability issue severely limits the throughput and responsiveness of large VAR models for high-resolution generation (e.g., $1024 \times 1024$ and beyond).

\noindent
\textbf{Training-Free Acceleration for Visual Generation.}
Training-free acceleration improves inference efficiency of pretrained models without additional fine-tuning by exploiting their intrinsic structure.
For diffusion models, such methods are well developed: improved samplers and scheduling~\cite{lu2025dpm,li2023autodiffusion,li2023snapfusion,jiang2025sada} reduce denoising iterations or adapt the trajectory, while feature reuse and caching~\cite{zou2024accelerating,ma2024deepcache,wimbauer2024cache,chen2024delta,zhang2024token,liu2025timestep,kahatapitiya2025adaptive} leverage slowly-varying activations across timesteps.
Chipmunk~\cite{chipmunk} further exploits activation sparsity by recomputing only the fastest-changing activations, showing that \emph{attention activation pattern reuse is a powerful training-free prior}.
In contrast, training-free acceleration for VAR remains underexplored.
FastVAR~\cite{fastvar} dynamically prunes tokens and skips the last high-resolution scales to obtain large speedups, but causes noticeable loss of high-frequency details and structural artifacts.
ScaleKV~\cite{scalekv} and HACK~\cite{hack} compress the KV cache and improve memory efficiency but yield limited runtime gains due to expensive key-value selection, while SkipVAR~\cite{skipvar} uses a lightweight discriminator for sample-adaptive scale skipping at the cost of retraining and quality degradation.
Our \NAME is a training-free acceleration framework that leverages VAR-specific attention activation sparsity for efficient computation, achieving over $\mathbf{1.5\times}$ speedup without skipping the last scales.

\section{Method}
\label{sec:method}

\subsection{Background}

Visual AutoRegressive (VAR) modeling~\cite{var} redefines the traditional autoregressive paradigm by shifting from a \textit{next-token} prediction to a \textit{next-scale} prediction paradigm. 
Instead of generating one token sequentially, each autoregressive step in VAR produces an entire token map at a specific resolution scale, progressively refining the image from coarse to fine.
Given an image feature map $\boldsymbol F \in \mathbb{R}^{H \times W \times D}$, VAR quantizes it into $K$ multi-scale residual token maps
\vspace{-2mm}
\begin{equation*}
    \mathcal{R} = \{\boldsymbol{R}_1, \boldsymbol{R}_2, \dots, \boldsymbol{R}_K\},
\vspace{-2mm}
\end{equation*}
where $\boldsymbol{R}_k$ denotes the token map at scale $k$. The resolution of $\boldsymbol{R}_k$ is $h_k\times w_k$ and it grows larger gradually from $k=1\to K$.
The size of last scale token map is $(h_K \times w_K)=(H\times W)$.
Based on the text prompt $t$ embedding $\boldsymbol\Psi(t)$ and $\mathcal{R}$, the joint distribution is factorized autoregressively as:
\vspace{-2mm}
\begin{equation*}
    p(\mathcal{R}) = \prod_{k=1}^{K} p(\boldsymbol{R}_k | \boldsymbol{R}_1,...,\boldsymbol{R}_{k-1}, \boldsymbol\Psi(t)),
    \label{eq:var_factor}
\vspace{-2mm}
\end{equation*}
where $\boldsymbol\Psi(t)$ typically is projected to the first scale with size $1\times1$.
During inference, the model predicts the next-scale residual $\boldsymbol{R}_k$ in parallel conditioned on all previous scales.

Let $N_k = h_k w_k$ denote the number of tokens at scale $k$, then the total number of tokens across all preceding scales including scale $k$ is given by $N_{\leq k} = \sum_{i=1}^{k} N_i$.
Each Transformer block employs causal attention, allowing tokens of scale $k$ to attend to tokens from all previous scales $\boldsymbol{R}_{\le k}$:
\vspace{-2mm}
\begin{equation}
    \mathbf{O}_k = \mathrm{Softmax}\left( \frac{\mathbf{Q}_k \mathbf{K}_{\le k}^{\top}}{\sqrt{d}} \right) \mathbf{V}_{\le k},
    \label{eq:attn_cross_scale}
    \vspace{-2mm}
\end{equation}
where $\mathbf{Q}_k\in \mathbb{R}^{B\times H\times N_k\times D}$ denotes the query projection from the current scale tokens, and $\mathbf{K}_{\le k}, \mathbf{V}_{\le k}\in \mathbb{R}^{B\times H\times N_{\le k}\times D}$ represent the concatenation of key and value projections from both the current and previous scales along the sequence dimension.
This enables coarse-to-fine dependency modeling, but as the scale spatial resolution increases, the number of tokens grows quadratically, resulting in $\mathcal{O}(n^4)$ attention complexity and substantial memory overhead.
Therefore, designing efficient cross-scale attention is crucial for scaling VAR inference to high resolutions---this motivates our \NAME.

\subsection{Cross-Scale Self-Similar Sparse Attention}

\noindent
\textbf{Motivation.}
%
To systematically analyze the pretrained VAR models~\cite{infinity,hart}, we visualize the attention maps at each scale and vertically concatenate them in ascending order of spatial resolution. 
The resulting visualization, shown in Fig.~\ref{fig:attn_map_3_feats}, clearly reveals how attention activation patterns evolve across scales.
From Fig.~\ref{fig:attn_map_3_feats}(b), it is evident that the attention activation patterns across different scales exhibit pronounced cross-scale similarity---the activation at scale $k$ closely mirrors that of the preceding scale $k-1$, particularly in the corresponding subregions when the attention map is partitioned by source scales.
Formally, for the full attention map $\mathbf{A}^{(k)} \in \mathbb{R}^{B\times H\times N_k \times N_{\le k}}$ at scale $k$, we divide it into $k$ column-blocks according to the source scales:
\vspace{-2mm}
\begin{equation}
    \mathbf{A}^{(k)} =
        \begin{bmatrix}
        \mathbf{A}^{(k,1)} &
        \mathbf{A}^{(k,2)} &
        \cdots &
        \mathbf{A}^{(k,k)}
        \end{bmatrix},
    \label{eq:attn_block}
    \vspace{-2mm}
\end{equation}
where each block $\mathbf{A}^{(k,i)} \in \mathbb{R}^{N_k \times N_i}$ corresponds to the attention from the current scale $k$ to the $i$-th previous scale ($i \le k$).
We empirically observe a strong similarity between consecutive scales:
\vspace{-2mm}
\begin{equation*}
    \mathbf{A}^{(k,i)} \approx \operatorname{Upsample}(\mathbf{A}^{(k-1,i-1)}), \quad \forall i \ge 2,
    \label{eq:cross_scale_sim}
    \vspace{-2mm}
\end{equation*}
suggesting that the activation pattern of the current scale can be predicted by upsampling that of the preceding scale.
This motivates us to reuse the sparse activation pattern identified at a mid-scale to predict sparse for larger scales, thereby avoiding redundant dense attention computations.

\noindent
\textbf{Sparse Decision Scale.}
We define the \textbf{sparse decision scale} $S$ as an optimal mid-scale whose dense attention yields a representative sparse pattern for predicting sparse activations of later scales.
Mathematically, $S$ is selected to minimize the discrepancy between its predicted sparse pattern (after cross-scale mapping) and the dense attention of the target high-resolution scale $k \in \{S+1, \dots,K\}$:
\vspace{-2mm}
\begin{equation*}
    S = \arg\min_{S \in \{1,\dots,K-1\}}
    \sum_{k=S+1}^{K}
    \mathcal{L}_{\text{perf}}\big(
    \mathcal{M}_{S\rightarrow k}(\Omega^{(S)}),
    \Omega^{(k)}_{\text{dense}}
    \big),
    \label{eq:sparse_decision_scale}
    \vspace{-2mm}
\end{equation*}
where $\Omega^{(S)}$ denotes the sparse activation set extracted from scale $S$ dense attention, 
$\mathcal{M}_{S\rightarrow k}(\cdot)$ is the index mapping function, 
and $\mathcal{L}_{\text{perf}}$ measures the generation fidelity gap induced by the predicted sparse pattern:
\vspace{-2mm}
\begin{equation*}
    \mathcal{L}_{\text{perf}} = - 10 \cdot \log_{10}\left( \frac{\text{MAX}_{\text{pixel\_value}}^2}{\operatorname{MSE}(\mathbf{I}_{\text{sparse}}, \mathbf{I}_{\text{base}})} \right),
    \vspace{-2mm}
\end{equation*}
where $\mathbf{I}_{\text{sparse}}=\mathbf{G}(\mathcal{M}_{S\rightarrow k}(\Omega^{(S)}))$ is the \NAME output using sparse patterns mapped from scale $S$.
Thus, $\mathcal{L}_{\text{perf}}$ is the negative value of PSNR, which measures the output's pixel-level differences between \NAME and baseline.
In practice, we fix $S$ to a mid-scale (e.g., scale 10), which, as shown in the ablations, achieves near-baseline generation performance while substantially reducing computational cost.
The sparse indices $\text{inds}^{(S)}$ extracted at this scale serve as the foundation for all subsequent sparse inference.

Specifically, at scale $S$, we compute full scaled dot-product attention following Eq.~\ref{eq:attn_cross_scale} (omitting $1/\sqrt{d}$):
\vspace{-2mm}
\begin{equation*}
    \mathbf{P}^{(S)} =
    \operatorname{Softmax}\left(
    \mathbf{Q}^{(S)}{\mathbf{K}^{(S)}}^{\top}
    \right), 
    \mathbf{O}^{(S)}_{\text{dense}} = \mathbf{P}^{(S)} \mathbf{V}^{(S)}.
    \vspace{-2mm}
\end{equation*}
We partition $\mathbf{P}^{(S)}$ along the query axis into $G_S = N_S / C$ contiguous query blocks of size $C$
\vspace{-2mm}
\begin{equation*}
    \mathbf{P}^{(S)} \in \mathbb{R}^{B\times H\times N_S \times N_{\le S}} \to \operatorname{reshape}(B, H, \frac{N_S}{C}, C, N_{\le S}),
    \vspace{-2mm}
\end{equation*}
and sum over each block to obtain the column-sum tensor~\cite{chipmunk}:
\vspace{-2mm}
\begin{equation*}
    \mathbf{D}^{(S)}_{b,h,i,j} = \sum_{q=ic}^{(i+1)c-1}\mathbf{P}^{(S)}_{b,h,q,j},
\end{equation*}
where $\mathbf{D}^{(S)} \in \mathbb{R}^{B\times H\times G_S\times N_{\le S}}$ encodes the total attention weight each block assigns to key $j$.
A top-k operation then identifies the most attended key indices:
\vspace{-1mm}
\begin{equation*}
    \text{inds}^{(S)}_{b,h,i,:\boldsymbol{k}} = \operatorname{TopK}_{\boldsymbol{k}}(\mathbf{D}^{(S)}_{b,h,i,:}).
    \vspace{-1mm}
\end{equation*}
Finally, we define the dense attention cache by subtracting the sparse component:
\vspace{-1mm}
\begin{equation*}
    \mathbf{O}^{(S)}_{\text{cache}} = \mathbf{O}^{(S)}_{\text{dense}} - \operatorname{Softmax}\left(\mathbf{Q}^{(S)}\mathbf{K}_{\text{inds}}^{(S)\top}
    \right)\mathbf{V}_{\text{inds}}^{(S)},
    \vspace{-1mm}
\end{equation*}
which captures the residual part of the dense output after removing the top sparse activations.

\noindent
\textbf{Cross-Scale Sparse Computation.}
For subsequent high-resolution scales $k > S$, we reuse the sparse indices $\text{inds}^{(S)}$ by mapping them across scales to obtain the target sparse indices $\text{inds}^{(k)}$.
We design an efficient \textbf{cross-scale sparse index mapping}, which projects $\text{inds}^{(S)}$ onto the spatial grid of scale $k$.
We simplify the computation of mapping as:
\begin{equation*}
    \text{inds}^{(k)} =
    \left\lfloor
    \frac{N_k}{N_S} \times \text{inds}^{(S)}
    \right\rfloor,
\end{equation*}
optionally adding a sink region $\mathcal{S}$ to guarantee coverage of global anchors: $\text{inds}^{(k)} \leftarrow \text{Concat}\big(\mathcal{S}, \text{inds}^{(k)}\big).$
This mapping preserves the hierarchical correspondence between scales, ensuring that attention sparsity remains coherent across resolutions.
Details are in the appendix.
The sparse update at scale $k$ is then efficiently computed as:
\vspace{-1mm}
\begin{equation*}
\begin{aligned}
    \Delta\mathbf{O}^{(k)} &=
    \operatorname{Softmax}\left(
    \mathbf{Q}^{(k)}
    {\mathbf{K}^{(k)\top}_{\text{inds}^{(k)}}}
    \right)
    \mathbf{V}^{(k)}_{\text{inds}^{(k)}}, \\
    \mathbf{O}^{(k)} &\approx \mathbf{O}^{(k)}_{\text{cache}} + \Delta\mathbf{O}^{(k)},
\end{aligned}
\vspace{-1mm}
\end{equation*}
where $\mathbf{O}^{(k)}_{\text{cache}}$ denotes the upsampled cache prediction from scale $S$.
As shown in Tab.~\ref{tab:ablation_sps-attn}, the $\mathbf{O}^{(k)}_\text{cache}$ term further enhances the high-frequency visual fidelity of the accelerated model with minimal latency.

\subsection{Cross-Scale Local Sparse Attention}

\begin{figure*}[ht]
    \centering
    \includegraphics[width=0.95\textwidth]{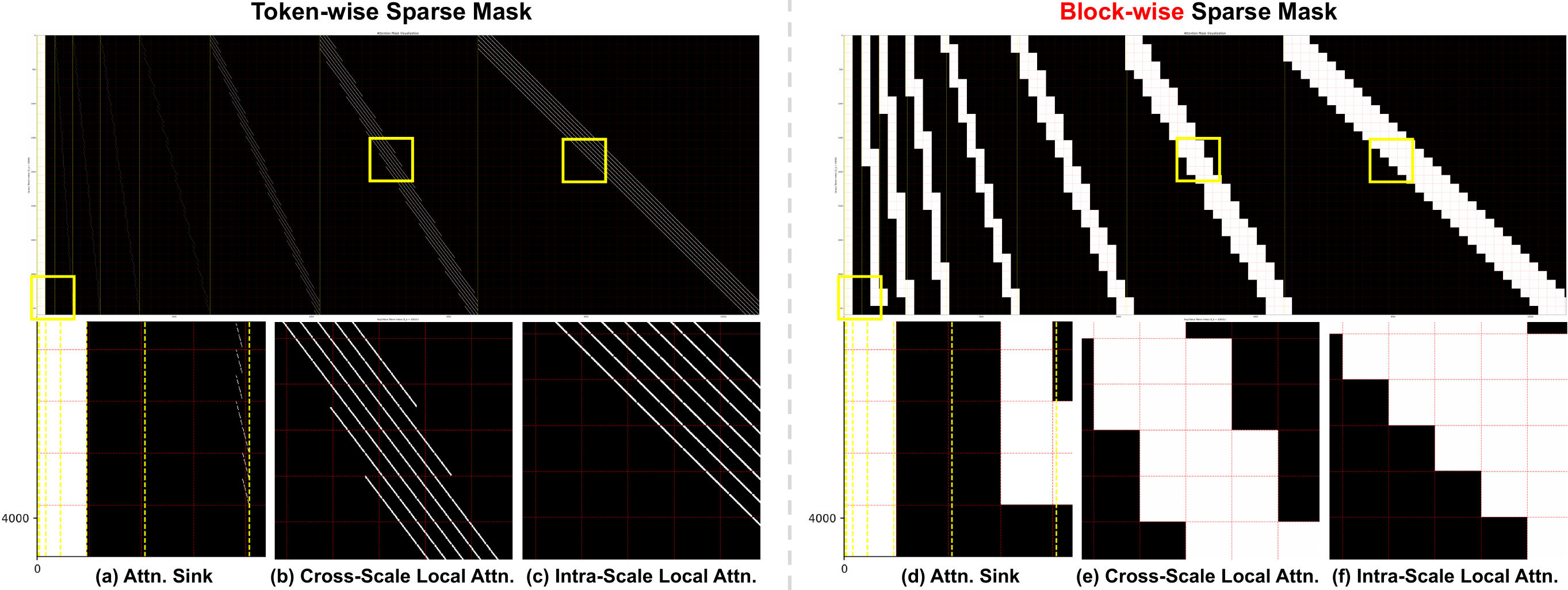}
    \vspace{-2mm}
    \caption{
        Visualization of the Cross-Scale Local Sparse Attention ($CSLA$) masks.
        This example shows the last scale attention map in Infinity ($q_{len}=4096$, $kv_{len}=10521$).
        The left shows the token-wise sparse mask, and the right shows the corresponding block-wise version after applying the block aggregation in Eq.~\ref{eq:block_mask}.
        Red dashed grids denote $128\times 128$ blocks, while yellow dashed lines mark the attention partition boundaries described in Eq.~\ref{eq:attn_block}.
        Zoom in for fine-detail visualization.
    }
    \label{fig:block-wise-sps}
    \vspace{-5mm}
\end{figure*}

\noindent 
\textbf{Motivation.}
Building upon the cross-scale sparse pattern reuse introduced in $CS^4A$, we further observe that in deep layers of pretrained VAR models, attention activation at later high-resolution scales primarily concentrates on two regions:
(i) \textbf{attention sink} formed by early scales, and 
(ii) \textbf{spatial locality} both within and across adjacent scales.
Hence, instead of relying on mapped sparse indices from decision scale $S$, directly exploiting these structural sparsity priors allows for greater acceleration benefits.
Moreover, recent advances~\cite{flashattention,flashattention2,flashattention3,triton,tilelang,flexattention,mva} in AI system for efficient operator optimization demonstrate that defining sparsity at the \textbf{block granularity} fully leverages modern GPU hardware features, achieving extreme operator acceleration.
Inspired by these insights, we propose \textbf{Cross-Scale Local Sparse Attention ($CSLA$)}, which constructs an efficient \textbf{block-wise sparse attention kernel} guided by sink and locality priors, further accelerating VAR inference while maintaining high-fidelity generation.

\noindent
\textbf{Cross-Scale Local Mapping.}
To exploit the observed spatial locality in VAR attention activation, we define a cross-scale local mapping function that restricts each query token to attend only to its nearby key–value tokens across scales.
For each query token $\mathbf{q}$ at scale $k$, we first identify its corresponding spatial location $(x_\mathbf{q}^{(k)}, y_\mathbf{q}^{(k)})$ and compute its aligned coordinates in the key grid of a historical scale $h \in \{1, ..., k-1, k\}$:
\vspace{-1mm}
\begin{equation*}
\scalebox{0.9}{$
    \left(\tilde{x}_\mathbf{q}^{(k\to h)}, \tilde{y}_\mathbf{q}^{(k\to h)}\right) = \left(\operatorname{round}\!\left(\frac{x_\mathbf{q}^{(k)}}{H_k} H_h\right),\operatorname{round}\!\left(\frac{y_\mathbf{q}^{(k)}}{W_k} W_h\right)\right),
$}
\vspace{-1mm}
\end{equation*}
where $(H_h, W_h)$ denotes the spatial resolution of historical scale $h$.
Each query thus attends to a local neighborhood centered at $(\tilde{x}_\mathbf{q}^{(k\to h)}, \tilde{y}_\mathbf{q}^{(k\to h)})$ with a per-scale window radius $r_h$, forming a token-wise local sparse mask
\vspace{-1mm}
\begin{equation*}
\scalebox{0.85}{$
    \mathbf{M}_{\text{local}}^{(k,h)}(\mathbf{q},\mathbf{k}) = \mathbf{1} \Big( |\tilde{x}_\mathbf{q}^{(k\to h)} - x_\mathbf{k}^{(h)}| \le r_h \ \wedge \ |\tilde{y}_\mathbf{q}^{(k\to h)} - y_\mathbf{k}^{(h)}| \le r_h \Big).
$}
\label{eq:wind_r}
\vspace{-1mm}
\end{equation*}
Intuitively, $\mathbf{M}_{\text{local}}^{(k,h)}(\mathbf{q},\mathbf{k})$ identifies whether key $\mathbf{k}$ lies within the local neighborhood of query $\mathbf{q}$ after rescaling between the two scales.
Meanwhile, a small number of early-scale tokens, denoted by the ``sink region'' $\mathcal{S}$, remain fully visible to all queries to preserve global context:
\vspace{-2mm}
\begin{equation*}
    \mathbf{M}_{\text{sink}}^{(k)}(\mathbf{q},\mathbf{k})=\mathbf{1}[\mathbf{k}\in\mathcal{S}].
    \vspace{-2mm}
\end{equation*}
Finally, we integrate the above local and sink masks into a unified \textbf{token-wise sparse mask}
\vspace{-3mm}
\begin{equation*}
    \mathbf{M}^{(k)} = \mathbf{M}_{\text{sink}}^{(k)} \vee \bigvee_{i=1}^{k} \mathbf{M}_{\text{local}}^{(k,h)},
    \vspace{-2mm}
\end{equation*}
where $\vee$ denotes the logical OR operation across multiple historical scales.

\noindent
\textbf{Block-wise Sparse Mask.}
To align with GPU-friendly block-sparse computations, we partition the query axis and the key axis uniformly into blocks of size $B$, where $u\in \{ 0, ..., \lceil N_k / B \rceil -1 \}$ denotes the block index along the query dimension and $v \in \{ 0, ..., \lceil N_{\le k} / B \rceil -1 \}$ represents the block index along the key dimension.
Furthermore, we construct an efficient block-wise sparse mask:
\vspace{-1mm}
\begin{equation}
    \mathbf{B}^{(k)}(u,v) = \bigvee_{\mathbf{q}\in\mathbf{B}_u} \bigvee_{\mathbf{k}\in\mathbf{B}_v} \mathbf{M}^{(k)}(\mathbf{q}, \mathbf{k}),
    \label{eq:block_mask}
    \vspace{-1mm}
\end{equation}
where each block groups $B\times B$ tokens.

Fig.~\ref{fig:block-wise-sps} provides an example of the sparse mask employed for attention computation at the last scale.
This conversion ensures that every active block represents a dense local window and inactive ones are completely pruned, producing a structured sparsity pattern well aligned with FlashAttention-style tiling.
We implement $CSLA$ based on top of the FlexAttention~\cite{flexattention} framework, with quantitative inference results summarized in Tab.~\ref{tab:kernel_speed}. Under a block size of 128, our $CSLA$ kernel achieves a $\mathbf{5.61\times}$ forward speed than FlashAttention on the last scale, and a remarkable $\mathbf{15.26\times}$ acceleration compared to the naïve token-wise sparse attention baseline.

\begin{table}[t]
    \centering
    \caption{
        Forward speed of sparse attention kernels in a setup aligned with Infinity-8B last scale inference configuration (bf16, $q\_len=4096$, $kv\_len=10521$, $d_{head}=128$, $heads=24$). Test on H100 GPU with CUDA 12.8 and PyTorch 2.7.1.
    }
    \vspace{-2mm}
    \resizebox{\linewidth}{!}{
    \begin{tabular}{lcccc}
        \toprule
        \textbf{Operation} & \textbf{Implementation} & \textbf{Sparsity (\%)} & \textbf{Latency (ms) $\downarrow$} & \textbf{Speedup $\uparrow$} \\
        \midrule
        FlashAttention2 & CUDA & 0.00\% & 3.0231 & $1.00\times$ \\
        F.sdpa + sparse mask & PyTorch & 87.86\% & 8.2276 & $0.37\times$ \\
        $CSLA$ (w/ block size 64) & FlexAttention & 83.50\% & 0.8396 & 3.60$\times$ \\
        $CSLA$ (w/ block size 128) & FlexAttention & 83.46\% & \textbf{0.5392} & \textbf{5.61$\times$} \\
        \bottomrule
    \end{tabular}
    \label{tab:kernel_speed}
    }
    \vspace{-4mm}
\end{table}

\section{Experiments}
\label{sec:exps}

\subsection{Experimental Setup}

\begin{table*}[htbp]
    \centering
    \caption{
        Quantitative comparison on efficiency and quality on 1024$\times$1024 GenEval~\cite{geneval} benchmarks.
        Note that the efficiency of Infinity baselines are tested under FlashAttention.
    }
    \vspace{-2mm}
  \resizebox{\textwidth}{!}{
    \begin{tabular}{lccccccccccc}
    \toprule
    \multirow{2}[4]{*}{\textbf{Methods}} & \multicolumn{4}{c}{\textbf{Acceleration}} & \multicolumn{4}{c}{\textbf{GenEval Score}} & \multicolumn{3}{c}{\textbf{Quality Metrics}} \\
\cmidrule(l){2-5}\cmidrule(l){6-9}\cmidrule(l){10-12}          & \#Scales $\downarrow$ & Speedup $\uparrow$ & Latency $\downarrow$ & Throughput $\uparrow$ & Two Obj. $\uparrow$ & Position $\uparrow$ & Color Attri. $\uparrow$ & Overall $\uparrow$ &  PSNR $\uparrow$ & SSIM $\uparrow$ & LPIPS $\downarrow$ \\
    \midrule
        & \multicolumn{11}{c}{\textit{w/o skip scales}} \\[2pt]
    Infinity-2B~\cite{infinity} & 13    & $1.00\times$ & 0.96s & 1.04it/s & \textbf{0.864} & 0.450 & 0.555 & 0.736 & -     & -     & - \\
    FastVAR~\cite{fastvar} & 13    & $1.01\times$ & 0.95s & 1.05it/s & 0.831 & 0.438 & 0.520 & 0.712 & 15.618 & 0.632 & 0.428 \\
    ScaleKV~\cite{scalekv} & 13    & $0.70\times$ & 1.37s & 0.73it/s & 0.854 & 0.453 & 0.548 & 0.732 & 23.370 & 0.820 & 0.192 \\
    \rowcolor{gray!15}
    \NAME (Ours) & 13    & $\mathbf{1.38\times}$ & \textbf{0.69s} & \textbf{1.44it/s} & 0.856 & \textbf{0.458} & \textbf{0.563} & \textbf{0.738} & \textbf{28.882} & \textbf{0.914} & \textbf{0.097} \\
    \hdashline[1pt/2pt]
    \noalign{\vskip 2pt}
          & \multicolumn{11}{c}{\textit{w/ skip last 2 scales}} \\[4pt]
    FastVAR~\cite{fastvar}    & 11    & $1.33\times$ & 0.72s & 1.39it/s & 0.831 & 0.405 & 0.575 & 0.716 & 12.867 & 0.544 & 0.538 \\
    \rowcolor{gray!15}
    \NAME (Ours)                & 11    & $\mathbf{1.70\times}$ & \textbf{0.56s} & \textbf{1.77it/s} & \textbf{0.846} & \textbf{0.413} & \textbf{0.588} & \textbf{0.725} & \textbf{15.272} & \textbf{0.596} & \textbf{0.474} \\
    \hdashline[1pt/2pt]
    \noalign{\vskip 2pt}
          & \multicolumn{11}{c}{\textit{w/ dynamic skip last scales}} \\[4pt]
    SkipVAR-2B~\cite{skipvar} & $10\sim13$ & $1.33\times$ & 0.72s & 1.39it/s & \textbf{0.854} & 0.418 & 0.565 & 0.730 & 17.650 & 0.678 & 0.391 \\
    \rowcolor{gray!15}
    \NAME (Ours)          & $10\sim13$    & $\mathbf{1.58\times}$ & \textbf{0.61s} & \textbf{1.65it/s} & \textbf{0.854} & \textbf{0.420} & \textbf{0.568} & \textbf{0.732} & \textbf{17.689} & \textbf{0.679} & \textbf{0.389} \\
    \midrule
          & \multicolumn{11}{c}{\textit{w/o skip scales}} \\[4pt]
    Infinity-8B~\cite{infinity} & 13    & $1.00\times$ & 1.65s & 0.61it/s & \textbf{0.899} & 0.610 & \textbf{0.695} & \textbf{0.798} & -     & -     & - \\
    FastVAR~\cite{fastvar} & 13    & $1.14\times$ & 1.45s & 0.69it/s & 0.917 & 0.610 & 0.680 & 0.792 & 17.403 & 0.630 & 0.333 \\
    ScaleKV~\cite{scalekv} & 13    & $0.67\times$ & 2.45s & 0.41it/s & 0.886 & 0.618 & 0.690 & 0.793 & 23.423 & 0.803 & 0.153 \\
    \rowcolor{gray!15}
    \NAME (Ours)             & 13    & $\mathbf{1.57\times}$ & \textbf{1.05s} & \textbf{0.95it/s} & 0.897 & \textbf{0.631} & 0.683 & 0.796 & \textbf{29.481} & \textbf{0.920} & \textbf{0.073} \\
    \hdashline[1pt/2pt]
    \noalign{\vskip 2pt}
          & \multicolumn{11}{c}{\textit{w/ skip last 2 scales}} \\[4pt]
    FastVAR~\cite{fastvar} & 11    & $1.79\times$ & 0.92s & 1.09it/s & 0.886 & 0.600 & 0.685 & 0.790 & 15.265 & 0.533 & 0.421 \\
    \rowcolor{gray!15}
    \NAME (Ours)             & 11    & $\mathbf{2.28\times}$ & \textbf{0.72s} & \textbf{1.38it/s} & \textbf{0.891} & \textbf{0.608} & \textbf{0.700} & \textbf{0.800} & \textbf{17.096} & \textbf{0.579} & \textbf{0.374} \\
    \hdashline[1pt/2pt]
    \noalign{\vskip 2pt}
          & \multicolumn{11}{c}{\textit{w/ dynamic skip last scales}} \\[4pt]
    SkipVAR-8B~\cite{skipvar} & $10\sim13$ & $1.71\times$ & 0.97s & 1.04it/s & 0.884 & 0.613 & 0.675 & 0.789 & 17.980 & 0.641 & 0.337 \\
    \rowcolor{gray!15}
    \NAME (Ours)          & $10\sim13$ & $\mathbf{2.05\times}$ & \textbf{0.80s} & \textbf{1.24it/s} & \textbf{0.894} & \textbf{0.625} & \textbf{0.695} & \textbf{0.796} & \textbf{19.614} & \textbf{0.694} & \textbf{0.296} \\
    \bottomrule
    \end{tabular}%
  }
  \label{tab:main_com}%
  \vspace{-2mm}
\end{table*}%

\noindent
\textbf{Base Models.}
We apply our proposed \textbf{\NAME} to the representative VAR-based text-to-image baselines, Infinity-2B, -8B~\cite{infinity} and HART~\cite{hart}, to verify the generality of our approach across different model and parameter scales.
For a fair comparison 
, all hyperparameters and model configurations are kept identical to the official Infinity implementations, except for the introduced sparse attention modules.
All our speed measurements and evaluation experiments are conducted on NVIDIA H100 GPUs.
For more implementation details, please refer to the appendix.

\noindent
\textbf{Evaluation Metrics.}
We evaluate the proposed method from two 
perspectives: generation quality and inference efficiency.
For generation quality, we consider both \textit{high-level} and \textit{low-level} metrics.
High-level evaluation focuses on semantic alignment and human preference, measured using 4 popular benchmarks---GenEval~\cite{geneval}, DPG-Bench~\cite{dpg} (in appendix),
ImageReward~\cite{imagereward} and HPSv2.1~\cite{hpsv2}.
%
For low-level evaluation, we take the baseline model’s outputs as the reference and employ PSNR, SSIM, and LPIPS to quantitatively assess the preservation of high-frequency textures and visual fidelity.
It is important to note that directly skipping late high-resolution scales can bring substantial acceleration to VAR models.
Therefore, we evaluate generation quality and inference efficiency under three settings:
(i) \textit{\textbf{without scale skipping}}, (ii) \textit{\textbf{skipping the last two scales}}~\cite{fastvar}, and (iii) \textit{\textbf{sample dynamic skipping}}~\cite{skipvar}.
This comprehensive and fair evaluation setup enables a more thorough comparison across different acceleration strategies.

\begin{table}[!ht]
    \centering
    \caption{
        Human preference evaluation on HPSv2.1~\cite{hpsv2} and ImageReward~\cite{imagereward}.
    }
    \vspace{-2mm}
    \resizebox{0.98\linewidth}{!}{
    \begin{tabular}{lcccccc}
        \toprule
        \multirow{2}{*}{\textbf{Methods}} &
        \multicolumn{3}{c}{\textbf{HPSv2.1}} &
        \multicolumn{1}{c}{\textbf{ImageReward}} \\
        \cmidrule(lr){2-4} \cmidrule(lr){5-5}
         & Photo $\uparrow$ & Concept-art $\uparrow$ & Average $\uparrow$ & Avg. Score $\uparrow$ \\
        \midrule
        \multicolumn{5}{c}{\textit{w/o skip scales}} \\
        Infinity-2B~\cite{infinity} & \textbf{29.45} & \textbf{30.47} & \textbf{30.55} & \textbf{0.9443} \\
        FastVAR~\cite{fastvar} & 28.59 & 29.81 & 29.85 & 0.9327 \\
        ScaleKV~\cite{scalekv} & 29.35 & 30.31 & 30.41 & 0.9327 \\
        \rowcolor{gray!15}
        \NAME (Ours) & 29.41 & 30.44 & 30.53 & 0.9416 \\
        \hdashline[1pt/2pt]
        \noalign{\vskip 2pt}
        \multicolumn{5}{c}{\textit{w/ skip last 2 scales}} \\
        FastVAR~\cite{fastvar} & 28.80 & 29.91 & 29.97 & 0.9191 \\
        \rowcolor{gray!15}
        \NAME (Ours) & \textbf{29.33} & \textbf{30.31} & \textbf{30.41} & \textbf{0.9201} \\
        \hdashline[1pt/2pt]
        \noalign{\vskip 2pt}
        \multicolumn{5}{c}{\textit{w/ dynamic skip last scales}} \\
        SkipVAR-2B~\cite{skipvar} & 29.31 & 30.38 & \textbf{30.47} & 0.9399 \\
        \rowcolor{gray!15}
        \NAME (Ours) & \textbf{29.33} & \textbf{30.39} & \textbf{30.47} & \textbf{0.9455} \\
        \midrule
        \multicolumn{5}{c}{\textit{w/o skip scales}} \\
        Infinity-8B~\cite{infinity} & \textbf{29.49} & 31.28 & 31.00 & 1.0529 \\
        FastVAR~\cite{fastvar} & 29.13 & 30.85 & 30.62 & 1.0380 \\
        ScaleKV~\cite{scalekv} & 29.37 & 31.10 & 30.82 & 1.0350 \\
        \rowcolor{gray!15}
        \NAME (Ours) & 29.47 & \textbf{31.34} & \textbf{31.01} & \textbf{1.0533} \\
        \hdashline[1pt/2pt]
        \noalign{\vskip 2pt}
        \multicolumn{5}{c}{\textit{w/ skip last 2 scales}} \\
        FastVAR~\cite{fastvar} & 28.82 & 30.38 & 30.23 & 1.0276 \\
        \NAME (Ours) & \textbf{29.19} & \textbf{30.84} & \textbf{30.59} & \textbf{1.0377} \\
        \hdashline[1pt/2pt]
        \noalign{\vskip 2pt}
        \multicolumn{5}{c}{\textit{w/ dynamic skip last scales}} \\
        SkipVAR-8B~\cite{skipvar} & 29.09 & 30.90 & 30.64 & 1.0297 \\
        \rowcolor{gray!15}
        \NAME (Ours) & \textbf{29.15} & \textbf{30.97} & \textbf{30.69} & \textbf{1.0305} \\
        \bottomrule
    \end{tabular}
    }
    \label{tab:human_pref}
\vspace{-6mm}
\end{table}

\subsection{Main Results}

\noindent
\textbf{Comparison on GenEval.}
We systematically evaluate \NAME on $1024\times 1024$ text-to-image generation, compared with state-of-the-art VAR acceleration methods in terms of acceleration, GenEval scores~\cite{geneval}, and low-level fidelity metrics.
The specific results are presented in Tab.~\ref{tab:main_com}.
Without skipping any scales, \NAME achieves $\mathbf{1.38\times}$ and $\mathbf{1.57\times}$ speedups on Infinity-2B and -8B respectively, while maintaining high-level semantic alignment nearly identical to the baseline.
Notably, it reduces the generation time of the \textbf{8B model} to the \textbf{1-second} regime for $\mathbf{1024\times 1024}$ generation.
Compared with FastVAR~\cite{fastvar} and ScaleKV~\cite{scalekv}, \NAME not only accelerates inference but also preserves high-frequency details more effectively, achieving substantially better low-level metrics.
Specifically, \NAME attains \textbf{PSNR approaching 30}, \textbf{SSIM above 0.9}, and \textbf{LPIPS below 0.1}, indicating that the proposed cross-scale sparse attention successfully reduces redundant computation while retaining the critical token interactions for fine-grained structure generation.
When combined with skip-scale strategies, \NAME further boosts acceleration to $\mathbf{1.70\times}$ and $\mathbf{2.28\times}$, increasing the inference throughput of the \textbf{2B model} to \textbf{1.77 it/s}.
%
Remarkably, even under the extreme acceleration scenario with scale skipping, \NAME still preserves high-frequency textures effectively, while maintaining leading low-level metrics.
%

\begin{figure}[!t]
    \centering
    \includegraphics[width=1\linewidth]{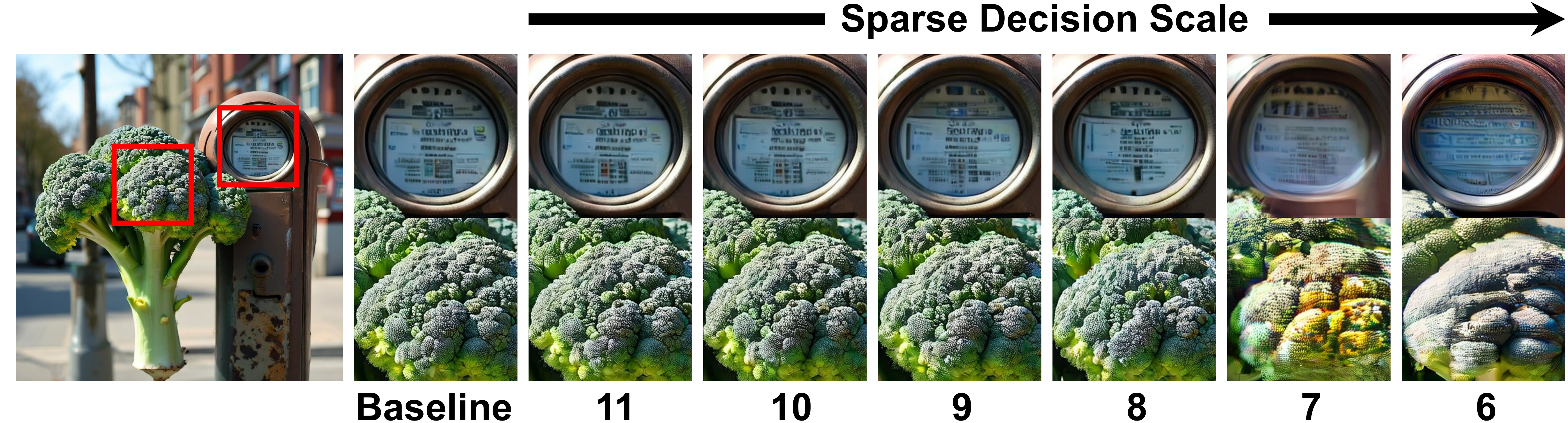}
    \vspace{-6mm}
    \caption{
        The impact of selecting sparse decision scales on generation results.
        Zoom in for fine-detail comparison.
    }
    \label{fig:decision_scale}
    \vspace{-6mm}
\end{figure}

\noindent
\textbf{Human Preference Evaluation.}
We further evaluate the generation quality of \NAME and other methods using human preference benchmarks~\cite{hpsv2,imagereward}.
As shown in Tab.~\ref{tab:human_pref}, both 2B and 8B model scales, \NAME without scale skipping achieves comparable human preference scores to the baseline, even attaining a score of \textbf{31.01} at 8B---\textbf{surpassing the baseline}.
This confirms our cross-scale sparse attention preserves fine semantic alignment and visual realism.
When combined with both scale-skipping setup, our \NAME still demonstrates remarkable robustness under aggressive acceleration.
We attribute this advantage to the suppression of attention noise through sparsity and the enhanced spatial locality that contributes to more consistent generation.
For additional qualitative results, please refer to the appendix.
%

\subsection{Ablation Studies}

\begin{figure}[t]
    \centering
    \includegraphics[width=1\linewidth]{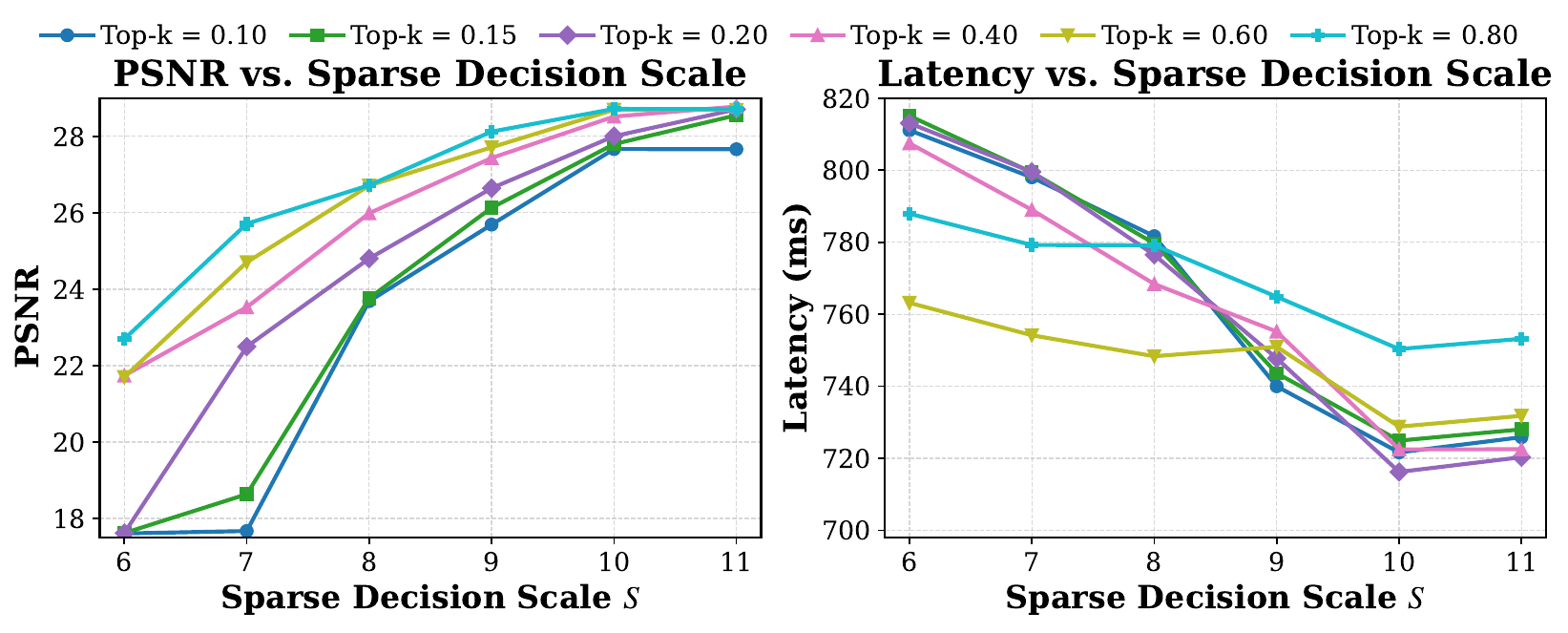}
    \vspace{-6mm}
    \caption{
        Ablation on sparse decision scale and Top-K.
    }
    \label{fig:ab_decision_scale}
    \vspace{-6mm}
\end{figure}

\noindent
\textbf{Sparse Decision Scale.}
To determine the optimal sparse decision scale, we conduct qualitative and quantitative ablations.
We first analyze how selecting different scales as the sparse decision scale affects final generation quality.
As shown in Fig.~\ref{fig:decision_scale}, using earlier scales (e.g., 6 or 7) introduces severe artifacts, texture corruption, and global blurring, whereas mid-to-late scales consistently preserve high-frequency details.
This behavior is tightly linked to the spatial resolution of VAR scales: earlier scales have limited resolution and exhibit coarse, highly dispersed attention patterns that share low similarity with later high-resolution sparse patterns.
Consequently, the cross-scale mapping becomes less accurate, leading to degraded generation quality.
We further perform systematic quantitative experiments to identify the best sparse decision scale and the optimal Top-K range.
As shown in Fig.~\ref{fig:ab_decision_scale}, scale 10 emerges as the best trade-off between quality and inference speed.
Across all settings, Top-K around 0.2 provides a stable and effective balance between sparsity and fidelity.

\noindent
\textbf{Local Sparsity.}
Tab.~\ref{tab:csla_ablation} ablates the attention sink range and window sizes configuration in $CSLA$.
We observe that retaining early-scale tokens (i.e., at least one dense sink block, shown as Fig.~\ref{fig:block-wise-sps}(d)) as attention sinks is crucial.
Without early-scale global structure, sparse attention at later large scales degrades generation quality, causing PSNR to drop from \textbf{25.897} to \textbf{23.541}.
%
For the last 3 scales, smaller windows yield the highest sparsity and fastest kernel latency, but degrade detail generation.
Balancing quality and efficiency, we adopt scale $\le 5$ as the default sinks and use window sizes [3, 5, 7] for the last 3 scales.

\begin{table}[t]
    \centering
    \caption{
        Ablation study on the attention sink and window size configuration in $CSLA$.
    }
    \vspace{-2mm}
    \resizebox{\linewidth}{!}{
    \begin{tabular}{@{}lccccc@{}}
        \toprule
        \textbf{Attn. Sink} & \textbf{Wind. Size} & \textbf{Latency (ms)} $\downarrow$ & \textbf{Sparsity} $\uparrow$ & \textbf{GenEval} $\uparrow$ & \textbf{PSNR} $\uparrow$ \\
        \midrule
        \multirow{7}{*}{Scale $\le$ 5}
         & [1, 3, 5]  & 0.4657 & 86.24\% & 0.727 & 25.657 \\
         & [3, 3, 3]  & 0.4480 & 86.77\% & 0.732 & 25.837 \\
         & [3, 5, 7]  & 0.5346 & 83.46\% & 0.730 & 25.897 \\
         & [5, 5, 5]  & 0.5266 & 83.77\% & 0.727 & 25.895 \\
         & [5, 7, 9]  & 0.6186 & 80.72\% & 0.729 & 25.944 \\
         & [7, 7, 7]  & 0.5960 & 81.18\% & 0.729 & 26.028 \\
         & [7, 9, 11] & 0.6889 & 78.06\% & 0.730 & 26.068 \\
        \midrule
        Scale $\le$ 6 & [3, 5, 7]  & 0.6114 & 81.03\% & 0.733 & 26.928 \\
        Scale $\le$ 7 & [3, 5, 7]  & 0.6717 & 78.60\% & 0.730 & 27.174 \\
        Scale $\le$ 8 & [3, 5, 7]  & 0.7729 & 75.21\% & 0.733 & 27.582 \\
        \textit{w/o} Sink      & [3, 5, 7]  & 0.5008 & 84.68\% & 0.714 & 23.541 \\
        \bottomrule
    \end{tabular}
    }
    \label{tab:csla_ablation}
\end{table}

\noindent
\textbf{Effectiveness of Cross-Scale Sparse Attention.}
We conduct ablations to analyze the effectiveness of our plug-and-play sparse attention modules.
As shown in Tab.~\ref{tab:ablation_sps-attn}, introducing $CS^4A$ alone yields a substantial acceleration of $\mathbf{1.46\times}$ while maintaining competitive generation quality.
A key design choice in $CS^4A$ is whether to incorporate the cached output $\mathbf{O}^{(k)}_{\text{cache}}$ from the sparse decision scale to supplement subsequent large-scale sparse outputs.
Results show that adding $\mathbf{O}^{(k)}_{\text{cache}}$ improves visual fidelity---increasing PSNR by approximately 0.7---with negligible latency overhead.
Finally, integrating the $CSLA$ further enhances spatial coherence 
, achieving the best overall performance with a $\mathbf{1.57\times}$ speedup and superior generation quality.
%
These results highlight the importance of intra- and inter-scale spatial locality in VAR attention, suggesting that leveraging such local priors provides a powerful basis for training-free acceleration.

\begin{table}[t]
    \centering
    \caption{
        Ablation study on our \textbf{plug-and-play efficient sparse attention modules}.
    }
    \vspace{-2mm}
    \resizebox{\linewidth}{!}{
    \begin{tabular}{lccccc}
        \toprule
        \textbf{Methods} & \textbf{Speedup} $\uparrow$ & \textbf{Latency} $\downarrow$ & \textbf{PSNR} $\uparrow$ & \textbf{SSIM} $\uparrow$ & \textbf{LPIPS} $\downarrow$ \\
        \midrule
        Infinity-8B~\cite{infinity} & 1.00$\times$ & 1.65s & - & - & - \\
        + $CS^4A$ (w/o $\mathbf{O}^{(k)}_{\text{cache}}$) & 1.46$\times$ & 1.13s & 25.665 & 0.837 & 0.131 \\
        + $CS^4A$ (w/ ~\ $\mathbf{O}^{(k)}_{\text{cache}}$)  & 1.43$\times$ & 1.15s & 26.359 & 0.860 & 0.114 \\
        + $CSLA$ (block-wise) & \textbf{1.57$\times$} & \textbf{1.05s} & \textbf{29.481} & \textbf{0.920} & \textbf{0.073} \\
        \bottomrule
    \end{tabular}
    }
    \label{tab:ablation_sps-attn}
    \vspace{-4mm}
\end{table}

\section{Conclusion}
\label{sec:conclusion}

We introduce \NAME, a training-free acceleration framework for VAR models that exploits the intrinsic sparsity of cross-scale attention.
By systematically analyzing pretrained VAR, we identify three key properties---\emph{attention sinks}, \emph{cross-scale activation similarity}, and \emph{spatial locality}---and use them to design two plug-and-play sparse attention modules:
Cross-Scale Self-Similar Sparse Attention ($CS^4A$), which predicts high-resolution scales sparse patterns from a sparse decision scale, and Cross-Scale Local Sparse Attention ($CSLA$), 
which enforces locality via a block-wise sparse kernel with a forward pass over 5$\times$ faster than FlashAttention.
On $1024\times1024$ text-to-image generation with Infinity-8B, \NAME achieves a $1.57\times$ speedup without skipping any scales while preserving almost all high-frequency details and human preference scores comparable to or better than the baseline.
When combined with existing scale-skipping strategies, \NAME further attains up to $2.28\times$ acceleration with competitive high-level and low-level metrics.
These results show that leveraging cross-scale attention sparsity is a principled and effective direction for efficient VAR inference.




\section{Acknowledgment}
This work was supported by the National Natural Science Foundation of China (NO.62572471).
We are also thankful to all the reviewers for their insightful comments and rigorous evaluation.

{
    \small
    \bibliographystyle{ieeenat_fullname}
    \bibliography{main}
}

\clearpage
\setcounter{page}{1}
\maketitlesupplementary

\appendix

\begin{figure*}[!ht]
    \centering
    \includegraphics[width=0.95\textwidth]{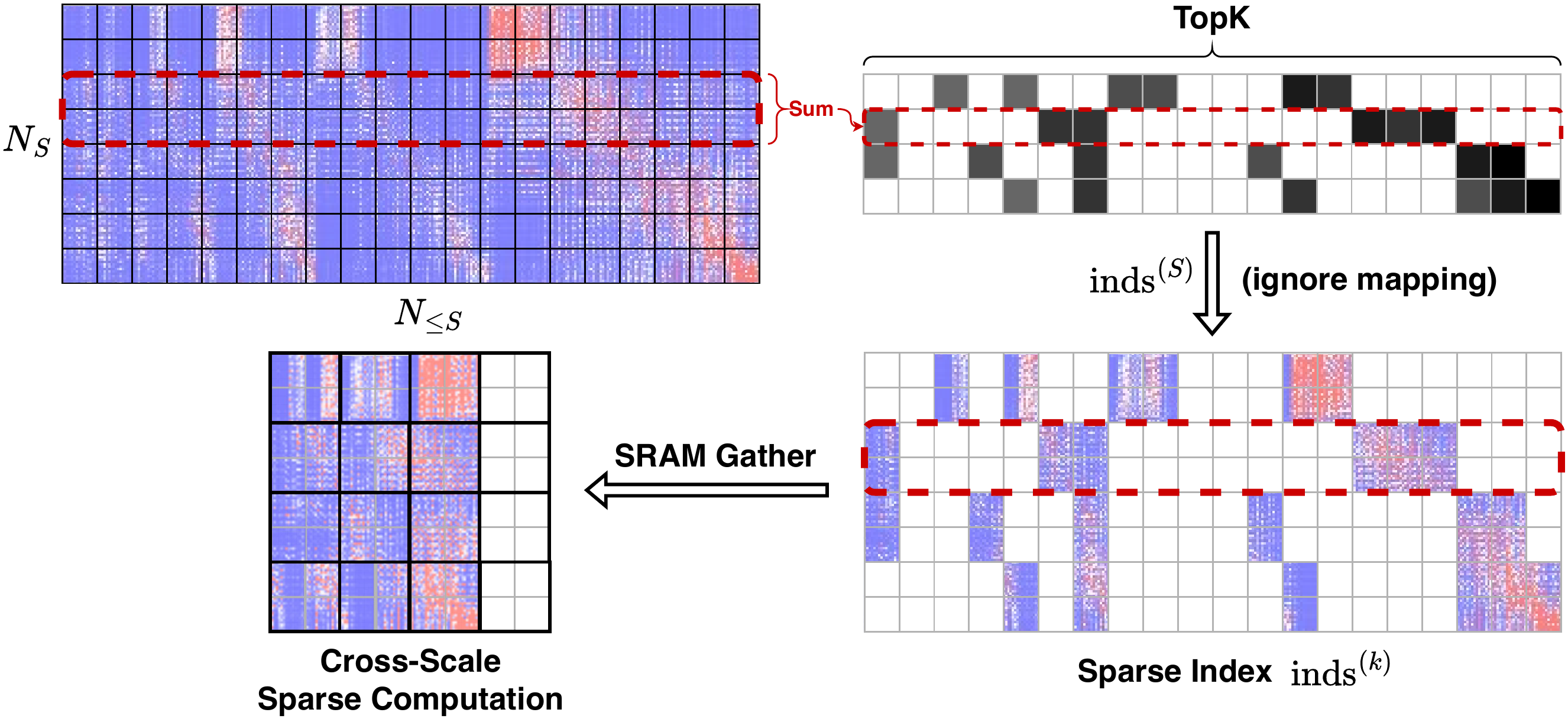}
    \caption{
        The illustration of $CS^4A$.
        The sparse index mapping process is omitted for clarity.
    }
    \label{fig:CS4A}
\end{figure*}

\begin{figure*}[ht]
    \centering
    \includegraphics[width=1\linewidth]{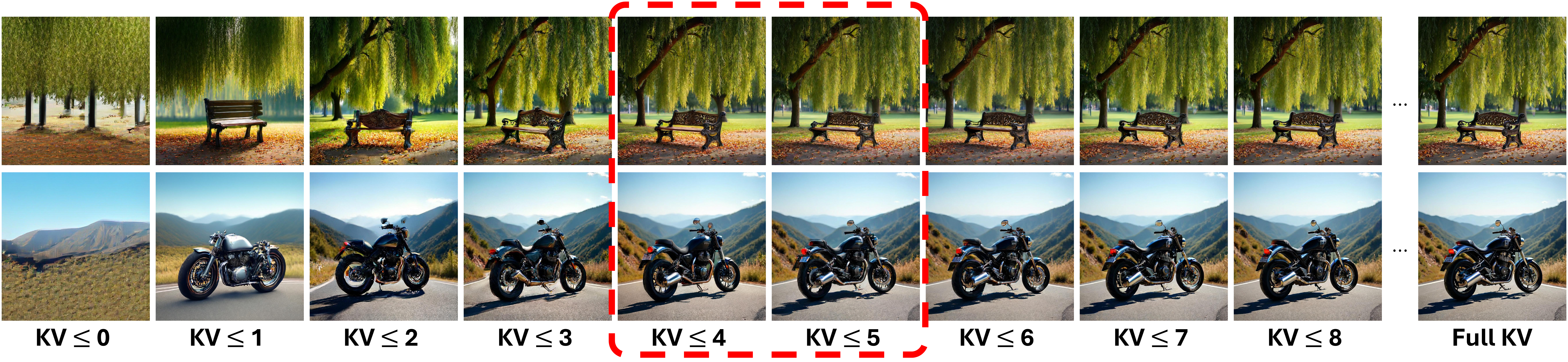}
    \caption{
        The influence of attention sinks in KV cache on generated images.
    }
   \label{fig:big_kv_sink}
\end{figure*}

\section{Algorithmic Details of $\boldsymbol{CS^4A}$}

\subsection{Hardware-Efficient Sparse Computation}
As illustrated in Fig.~\ref{fig:CS4A}, the pipeline of $CS^4A$ comprises two distinct phases: \textbf{sparse pattern identification} and \textbf{hardware-accelerated sparse computation}.
At the sparse decision scale $S$, we compute full dense attention to capture the global dependency structure.
To avoid the computational overhead of sorting per-query token attention scores, we adopt the block-wise column-sum strategy~\cite{chipmunk}.
We partition the attention map $\mathbf{P}^{(S)}$ into contiguous query blocks of size $C$, highlighted by the red dashed boxes in Fig.~\ref{fig:CS4A}.
Within each query block, we aggregate attention scores along the query dimension to derive a cumulative importance score for each key column.
Subsequently, the sparse indices $\text{inds}^{(S)}$ are determined by performing a Top-$K$ selection on these column sums.
These indices are then projected to the target scale $k \in \{S+1, \dots, K\}$ via our sparse index mapping strategy $\mathcal{M}_{S\to k}(\cdot)$, enabling efficient sparse computation at high-resolution scales.

A primary challenge in sparse attention lies in the inefficiency of unstructured sparsity, where random sparse matrix multiplication fails to effectively leverage the Tensor Cores on modern GPUs. Furthermore, runtime overheads such as dynamic mask computation and cache maintenance can introduce significant latency.
To address this, we adopt the Tile Packing technique~\cite{chipmunk}, which maps sparse computation into compacted dense computation. We extend their efficient custom kernels to support our causal cross-scale attention mechanism.
Specifically, guided by the mapped sparse indices $\text{inds}^{(k)}$, the kernel \textbf{gathers} discrete, non-contiguous key and value vectors from the GPU global memory (HBM).
These vectors are then \textbf{packed} into contiguous dense tiles within the GPU shared memory (SRAM).
Once resident in SRAM, these packed tiles constitute a logically dense matrix (as shown in the bottom-left of Fig.~\ref{fig:CS4A}). 
This transformation enables the GPU to execute standard dense General Matrix Multiplications (GEMMs) using Tensor Cores, thereby achieving peak computational throughput.


\subsection{Cross-Scale Sparse Index Mapping}
To efficiently propagate the sparse activation patterns from the decision scale $S$ to a higher-resolution target scale $k \in \{S+1, \dots, K\}$, we propose a two-stage mapping mechanism, denoted as $\mathcal{M}_{S\to k}(\cdot)$. 
This mapping addresses two fundamental challenges: 
the quadratic expansion of the query grid and 
the scale drift of historical key-value pairs. 
Specifically, we decompose $\mathcal{M}$ into two coupled transformations: \textit{Query Block Homography} and \textit{Relative Scale Alignment}.

\paragraph{Query Block Homography.}
Following the hardware-efficient Column-Sum design~\cite{chipmunk}, query tokens are partitioned into contiguous blocks of size $C$ (e.g., $C=192$).
As the spatial resolution expands from $h_S\times w_S$ to $h_k\times w_k$, the number of query blocks increases from $G_S = \lceil N_S / C \rceil$ to $G_k = \lceil N_k / C \rceil$.
Let $\mathcal{G}_{S}=\{0, \dots, G_S-1\}$ and $\mathcal{G}_k=\{0, \dots, G_{k-1}\}$ denote the sets of query block indices at scales $S$ and $k$, respectively.
We model the mapping from a target block $g_k \in \mathcal{G}_{k}$ to a source block $g_S \in \mathcal{G}_S$ as a nearest-neighbor interpolation within the normalized coordinate space $[0, 1]$.
Formally, the mapping is defined as:
\begin{equation*}
    \phi(g_k) = \left\lfloor \frac{g_k + 0.5}{G_k} \cdot G_S - 0.5 \right\rceil_{\mathcal{G}_S},
\end{equation*}
where $\lfloor \cdot \rceil_{\mathcal{G}_S}$ denotes rounding to the nearest integer clipped within the bounds of $\mathcal{G}_{S}$.
This transformation ensures that query tokens at the finer scale $k$ inherit the sparsity pattern from their spatially corresponding region at the coarser scale $S$, thereby preserving the global attention structure.

\paragraph{Relative Scale Alignment for KVs.}
The KV cache in VAR is a concatenation of tokens from all historical scales, e.g., for the sparse decision scale $S$, its KV cache $\mathbf{K}_{\le S} = \operatorname{Concat}(\mathbf{K}_1, \dots, \mathbf{K}_S)$.
A flattened global KV index $j$ is structurally ambiguous without scale context, i.e., $j$ does not explicitly encode which historical scale it comes from, making direct cross-scale mapping ambiguous.
%
To resolve this ambiguity, we propose a \textit{Decompose--Align--Project} mechanism to map sparse KV indices from scale $S$ to later target scales $k \in \{S+1, \dots, K\}$.

\begin{itemize}[leftmargin=*]
    \item \textbf{Decomposition.}
    We first decompose each global KV index $j$ at scale $S$ into its source scale and its within-scale offset.
    Let
    \vspace{-3mm}
    \begin{equation*}
        C_0 = 0, \qquad
        C_i = \sum_{r=1}^{i} N_r, \quad i=1,\dots,S,
        \vspace{-1mm}
    \end{equation*}
    where $N_i$ is the number of tokens at scale $i$.
    Then the source scale $l$ of index $j$ is determined by
    \begin{equation*}
        C_{l-1} \le j < C_l,
    \end{equation*}
    and the corresponding local offset is
    \begin{equation*}
        \delta = j - C_{l-1}.
    \end{equation*}
    Thus, each flattened index is decomposed into a tuple $(l,\delta)$, where $l$ identifies the historical source scale and $\delta$ specifies the relative position within that scale.

    \item \textbf{Relative Alignment.}
    Based on the cross-scale self-similarity observation, the sparse attention patterns depend on the \textit{relative scale distance} between the current query scale and the historical key scales.
    Hence, when mapping from the sparse decision scale $S$ to a target scale $k$, we preserve this relative offset.
    Specifically, if a KV index at scale $S$ comes from historical scale $l$, then its aligned source scale at target scale $k$ is
    \begin{equation*}
        l' = k - (S - l).
        \label{eq:relative_alignment}
    \end{equation*}
    Equivalently, this enforces
    \begin{equation*}
        S-l = k-l',
    \end{equation*}
    so that the mapped interaction at scale $k$ corresponds to the same relative cross-scale region as at scale $S$.
    For example, the sparse pattern associated with $\mathbf{A}^{(S,l)}$ is mapped to the corresponding region in $\mathbf{A}^{(k,l')}$ with identical relative distance.

    \item \textbf{Spatial Projection.}
    Finally, we project the local spatial offset $\delta$ from the resolution of scale $l$ ($h_l \times w_l$) to the resolution of the target scale $l'$ ($h_{l'} \times w_{l'}$).
    To preserve 2D spatial locality, we de-linearize $\delta$ into 2D coordinates $(u, v)$, perform bilinear scaling, and re-linearize:
    \begin{equation*}
        u' = \lfloor u \cdot \frac{h_{l'}}{h_l} \rfloor,
        \quad
        v' = \lfloor v \cdot \frac{w_{l'}}{w_l} \rfloor,
        \quad
        \delta’ = u' \cdot w_{l'} + v'.
    \end{equation*}
    Finally, the mapped global KV index at target scale $k$ is reconstructed as
    \begin{equation*}
        j' = C_{l'-1} + \delta',
    \end{equation*}
    where $C_{l'-1}$ denotes the starting index of scale $l'$ in the KV cache of target scale $k$.
\end{itemize}

\begin{figure*}[!ht]
    \centering
    \includegraphics[width=1\textwidth]{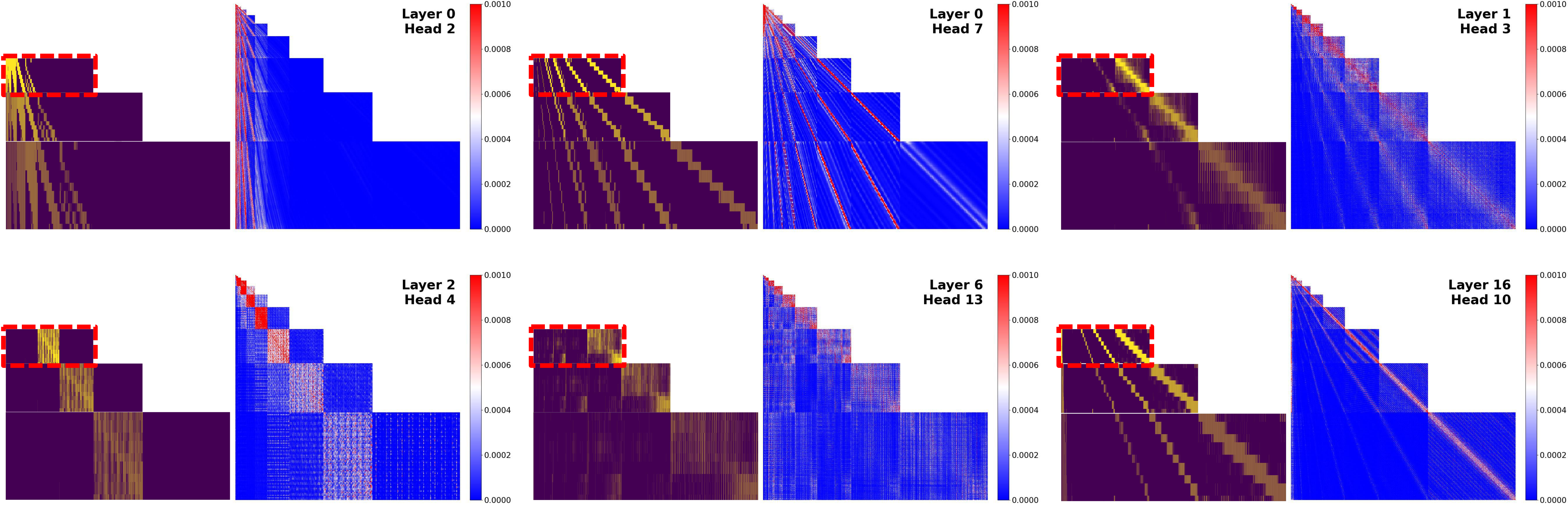}
    \caption{
        Visualization of cross-scale sparse index mapping.
        The \textcolor{red}{red dashed box} highlights the sparse indices identified at the sparse decision scale $S$, while the blocks below display the corresponding sparse indices projected onto subsequent high-resolution scales.
        Evidently, the mapped sparse indices precisely cover the most salient attention activation regions of the baseline model.
    }
    \label{fig:sps_index_map}
\end{figure*}

\noindent
This procedure preserves both relative cross-scale structure and within-scale locality, enabling sparse patterns identified at the sparse decision scale to be consistently reused at later high-resolution scales.

\paragraph{Attention Sink.}
To mitigate the loss of critical global context during sparse selection, we explicitly enforce a \textbf{sink token preservation} strategy.
Let $\mathcal{A}_{\text{sink}} = \{0, \dots, N_{\text{sink}}-1\}$ denote the set of indices corresponding to tokens from the initial scales.
The final sparse index set for the target scale $K$ is constructed as the union of the mapped dynamic indices and these static sink indices:
\begin{equation*}
    \text{inds}^{(k)} = \mathcal{A}_{\text{sink}} \cup \mathcal{M}_{S \to k}(\text{inds}^{(S)}).
\end{equation*}
This ensures that the accelerated sparse attention mechanism retains a stable attention sink, thereby preventing the collapse of semantic coherence during the generation process.

\paragraph{Effectiveness Analysis.}
To validate the effectiveness of the proposed sparse index mapping strategy, we visualize the alignment between the predicted sparse patterns and the full attention activations of the baseline model in Fig.~\ref{fig:sps_index_map}.
Specifically, we visualize the sparse indices derived from the sparse decision scale $S$ (highlighted by red dashed boxes) alongside the mapped indices at subsequent high-resolution scales, allowing for a direct examination of the sparse patterns predicted by $CS^4A$.
As illustrated across various layers and heads, the mapped sparse indices (depicted in gold) exhibit a remarkable spatial correspondence with the ground-truth activation hotspots (red regions) of the baseline model.
Notably, this alignment persists even in the final scales where the resolution expands significantly, confirming that the structural self-similarity of cross-scale attention is robustly preserved by our mapping strategy.
This coverage of salient attention activation regions ensures that the sparse computation approximates the full attention mechanism with minimal error.
It fundamentally explains why $CS^4A$ achieves substantial acceleration while maintaining the generation of high-frequency details, as the model continues to attend to the most critical contextual information during the fine-grained refinement stages.

\begin{table*}[ht]
    \centering
    \caption{
        Quantitative comparison on the DPG-Bench dataset. 
        The best results in each setting are highlighted in \textbf{bold}.
    }
    \label{tab:dpg_results}
    \resizebox{0.9\textwidth}{!}{%
    \begin{tabular}{lcccccccc}
        \toprule
        \multirow{2}{*}{\textbf{Methods}} & \multicolumn{2}{c}{\textbf{Acceleration}} & \multicolumn{6}{c}{\textbf{DPG-Bench}} \\
        \cmidrule(lr){2-3} \cmidrule(l){4-9}
         & \#Scales $\downarrow$ & Speedup $\uparrow$ & Global $\uparrow$ & Entity $\uparrow$ & Attribute $\uparrow$ & Relation $\uparrow$ & Other $\uparrow$ & Overall $\uparrow$ \\
        \midrule
        \multicolumn{9}{c}{\textit{w/o skip scales}} \\
        Infinity-2B~\cite{infinity} & 13 & $1.00\times$ & 88.499 & 88.120 & \textbf{88.957} & 88.767 & 84.521 & 82.839 \\
        FastVAR~\cite{fastvar} & 13 & $1.01\times$ & 77.291 & 89.168 & 88.537 & \textbf{91.932} & 84.637 & 82.723 \\
        ScaleKV~\cite{scalekv} & 13 & $0.70\times$ & 78.088 & 88.550 & 87.495 & 90.229 & \textbf{89.468} & \textbf{82.885} \\
        \rowcolor{gray!15}
        \NAME (Ours) & 13 & $\mathbf{1.38\times}$ & \textbf{89.021} & \textbf{89.174} & 87.274 & 91.015 & 84.405 & 82.858 \\
        \hdashline[1pt/2pt]
        \noalign{\vskip 2pt}
        \multicolumn{9}{c}{\textit{w/ skip last 2 scales}} \\
        FastVAR~\cite{fastvar} & 11 & $1.33\times$ & 81.423 & 88.267 & 88.184 & 90.127 & \textbf{89.351} & 82.651 \\
        \rowcolor{gray!15}
        \NAME (Ours) & 11 & $\mathbf{1.70\times}$ & \textbf{83.864} & \textbf{88.945} & \textbf{88.274} & \textbf{90.883} & 81.888 & \textbf{82.663} \\
        \hdashline[1pt/2pt]
        \noalign{\vskip 2pt}
        \multicolumn{9}{c}{\textit{w/ dynamic skip last scales}} \\
        SkipVAR-2B~\cite{skipvar} & 10$\sim$13 & $1.33\times$ & 85.273 & \textbf{89.115} & \textbf{89.289} & \textbf{89.519} & 84.836 & 82.877 \\
        \rowcolor{gray!15}
        \NAME (Ours) & 10$\sim$13 & $\mathbf{1.58\times}$ & \textbf{89.216} & 88.896 & 88.004 & 89.279 & \textbf{89.366} & \textbf{82.891} \\
        \midrule
        \multicolumn{9}{c}{\textit{w/o skip scales}} \\
        Infinity-8B~\cite{infinity} & 13 & $1.00\times$ & \textbf{92.144} & 89.741 & 90.376 & 92.084 & 91.862 & \textbf{86.440} \\
        FastVAR~\cite{fastvar} & 13 & $1.14\times$ & 81.784 & 90.195 & 89.311 & 94.084 & 84.496 & 86.343 \\
        ScaleKV~\cite{scalekv} & 13 & $0.67\times$ & 81.876 & 90.041 & 89.666 & \textbf{94.464} & \textbf{92.675} & 86.333 \\
        \rowcolor{gray!15}
        \NAME (Ours) & 13 & $\mathbf{1.57}\times$ & 87.306 & \textbf{91.510} & \textbf{90.862} & 91.565 & 90.193 & 86.411 \\
        \hdashline[1pt/2pt]
        \noalign{\vskip 2pt}
        \multicolumn{9}{c}{\textit{w/ skip last 2 scales}} \\
        FastVAR~\cite{fastvar} & 11 & $1.79\times$ & 87.295 & \textbf{91.466} & 90.048 & \textbf{91.113} & \textbf{90.308} & 86.332 \\
        \rowcolor{gray!15}
        \NAME (Ours) & 11 & $\mathbf{2.28\times}$ & \textbf{91.729} & 91.068 & \textbf{91.031} & 90.354 & 89.072 & \textbf{86.468} \\
        \hdashline[1pt/2pt]
        \noalign{\vskip 2pt}
        \multicolumn{9}{c}{\textit{w/ dynamic skip last scales}} \\
        SkipVAR-8B~\cite{skipvar} & 10$\sim$13 & $1.71\times$ & \textbf{92.648} & 90.591 & \textbf{91.032} & 90.088 & 86.008 & 86.293 \\
        \rowcolor{gray!15}
        \NAME (Ours) & 10$\sim$13 & $\mathbf{2.05\times}$ & 88.227 & \textbf{91.863} & 90.703 & \textbf{92.239} & \textbf{86.758} & \textbf{86.409} \\
        \bottomrule
    \end{tabular}%
    }
\end{table*}

\begin{table*}[ht]
    \centering
    \caption{
        Quantitative comparison of memory consumption.
        ``Torch Mem.'' refers to peak PyTorch allocated memory, and ``NV Mem.'' refers to NVIDIA physical memory.
        The best results in each setting are highlighted in \textbf{bold}.
    }
    \label{tab:efficiency_memory}
    \resizebox{\textwidth}{!}{%
    \begin{tabular}{lcccccccc}
        \toprule
        \multirow{2}{*}{\textbf{Methods}} & \multicolumn{4}{c}{\textbf{2B model}} & \multicolumn{4}{c}{\textbf{8B model}} \\
        \cmidrule(lr){2-5} \cmidrule(lr){6-9}
         & GenEval & Speedup $\uparrow$ & \textbf{Torch Mem.} $\downarrow$ & \textbf{NV Mem.} $\downarrow$ & GenEval $\uparrow$ & Speedup $\uparrow$ & \textbf{Torch Mem.} $\downarrow$ & \textbf{NV Mem.} $\downarrow$ \\
        \midrule
        \multicolumn{9}{c}{\textit{w/o skip scales}} \\[2pt]
        Infinity~\cite{infinity} & 0.736 & 1.00$\times$ & 15.78GB & 28.63GB & \textbf{0.798} & 1.00$\times$ & 37.11GB & 55.75GB \\
        FastVAR~\cite{fastvar} & 0.712 & 1.01$\times$ & 15.91GB & 28.45GB & 0.792 & 1.14$\times$ & 37.41GB & 54.29GB \\
        ScaleKV~\cite{scalekv} & 0.732 & 0.70$\times$ & \textbf{10.98GB} & \textbf{17.01GB} & 0.793 & 0.67$\times$ & \textbf{22.71GB} & \textbf{29.64GB} \\
        \rowcolor{gray!15}
        \NAME (Ours) & \textbf{0.738} & \textbf{1.38$\times$} & 16.02GB & 23.21GB & 0.796 & \textbf{1.57$\times$} & 37.48GB & 45.45GB \\
        \rowcolor{gray!15}
        \ \ + KV comp. & 0.730 & 1.29$\times$ & 12.74GB & 18.51GB & 0.793 & 1.31$\times$ & 30.42GB & 35.18GB \\
        \hdashline[1pt/2pt]
        \noalign{\vskip 2pt}
        
        \multicolumn{9}{c}{\textit{w/ skip last 2 scales}} \\
        FastVAR~\cite{fastvar} & 0.716 & 1.33$\times$ & 11.65GB & 19.73GB & 0.79 & 1.79$\times$ & \textbf{26.19GB} & 37.31GB \\
        \rowcolor{gray!15}
        \NAME (Ours) & \textbf{0.725} & \textbf{1.70$\times$} & \textbf{11.12GB} & \textbf{17.04GB} & \textbf{0.800} & \textbf{2.28$\times$} & 26.42GB & \textbf{29.73GB} \\
        \hdashline[1pt/2pt]
        \noalign{\vskip 2pt}
        
        \multicolumn{9}{c}{\textit{w/ dynamic skip last scales}} \\
        SkipVAR~\cite{skipvar} & 0.730 & 1.33$\times$ & \textbf{12.50GB} & 23.32GB & 0.789 & 1.71$\times$ & 30.05GB & 44.00GB \\
        \rowcolor{gray!15}
        \NAME (Ours) & \textbf{0.732} & \textbf{1.58$\times$} & \textbf{12.50GB} & \textbf{16.98GB} & \textbf{0.796} & \textbf{2.05$\times$} & \textbf{30.00GB} & \textbf{34.10GB} \\
        \bottomrule
    \end{tabular}%
    }
\end{table*}

\subsection{Difference from Chipmunk.}
Chipmunk achieves training-free acceleration by exploiting the temporal redundancy and the similarity of attention activations across denoising steps inherent in Diffusion Transformers (DiTs)~\cite{flux_kontext,hunyuanvideo,wan}.
However, the premise that latent variables maintain a constant spatial resolution during denoising does not hold for the unique next-scale prediction paradigm of VAR. 
Fundamentally, VAR is characterized by a progressive increase in token count as spatial resolution grows during inference, standing in stark contrast to DiT, where the sequence length remains invariant.
The motivation behind our proposed $CS^4A$ lies in the cross-scale self-similarity intrinsic to VAR models, namely that attention activation patterns at smaller scales exhibit similarities to those at larger scales in corresponding regions.
To the best of our knowledge, \NAME is the first work to identify and exploit this attention activation pattern redundancy for accelerating visual autoregressive generation, exploring a dimension of sparsity orthogonal to the temporal sparsity found in DiTs.
Furthermore, to effectively propagate this sparsity across scales, we propose an efficient sparse index mapping strategy that projects sparse patterns from low-resolution to high-resolution scales via rigorous geometric transformations.
From an engineering perspective, we extend Chipmunk's highly optimized sparse kernels---originally designed solely for standard self-attention---to support the causal cross-scale attention required by VAR. 
This generalization unlocks the potential of sparse computing for a broader class of autoregressive models beyond the fixed-length setting of DiTs.

\section{Complexity of Cross-Scale Sparse Attention}
We analyze the time complexity of the proposed $CS^4A$ and $CSLA$, comparing them against standard cross-scale full attention.
Let $N_k$ denote the number of tokens at the current scale $k$, and $N_{\le k}$ represent the cumulative number of tokens across all preceding scales. The feature dimension is denoted by $d$.
For a given scale $k$, following Eq.~\ref{eq:attn_cross_scale}, the computational complexity of full dense attention is given by:
\begin{equation*}
    \mathcal{O}(d N_k N_{\le k}).
\end{equation*}

\paragraph{$\boldsymbol{CS^4A}$.}
$CS^4A$ initially performs full dense attention at the sparse decision scale $S$ to identify salient sparse patterns.
The complexity of full attention at scale $S$ is $\mathcal{O}(d N_S N_{\le S})$, consistent with the baseline model. 
Since $S$ typically corresponds to a mid-resolution scale, this computational overhead is relatively marginal.
Additionally, computing the column-sum tensor $D^{(S)}$ per query block introduces a complexity of $\mathcal{O}(d G_S N_{\le S})$, where $G_S = \left \lceil N_S / C \right \rceil$.
Consequently, the total complexity at the sparse decision scale $S$ is given by:
\begin{equation*}
    \mathcal{O}^{(S)} = \mathcal{O}(d N_S N_{\le S}) + \mathcal{O}(d G_S N_{\le S}).
\end{equation*}
In practice, hardware-friendly block size $C$ (e.g., 128 or 192) makes the second term negligible compared to the first.

For subsequent high-resolution scales $k > S$, we employ cross-scale sparse computation, which restricts attention calculation to the Top-K selected KV pairs.
The process begins with sparse index mapping. Mapping the query blocks and KV indices incurs a time complexity of $\mathcal{O}(N_k)$, as it involves only lightweight coordinate transformations and gather operations.
Let $\alpha$ denote the sparsity ratio (e.g., retaining the top 20\% of KVs).
For any given query, the number of active KVs is $\alpha \cdot N_{\le k} \ll N_{\le k}$.
Therefore, the complexity of the sparse computation is reduced to 
\begin{equation*}
    \mathcal{O}(d \alpha N_k N_{\le k}).
\end{equation*}
This implies a linear relationship between total complexity and the sparsity ratio.
For small $\alpha$, $CS^4A$ achieves significant acceleration. Theoretically, as $N_k$ increases, the complexity reduction factor asymptotically approaches $1/\alpha$.

\paragraph{$\boldsymbol{CSLA}$.}
$CSLA$ decomposes the historical KV cache into a dense attention sink region and a block-wise local sparse region.
The first $N_{\text{sink}}$ KV tokens from early scales are always attended with full dense attention, preserving their role as global structural anchors~\cite{streamingllm}.
The remaining $N_{\le k} - N_{\text{sink}}$ KV tokens are arranged as 2D grids per historical scale, where each scale $h \in \{1,\dots,k\}$ is assigned a scale-dependent local window size $\text{w}_h$ (with radius $r_h = \left \lfloor w_h / 2 \right \rfloor$).
%
Consequently, for any query at scale $k$, the number of KV tokens involved in the sparse region is upper-bounded by
\begin{equation*}
    N_{\text{local}} \le \sum_{h\le k}\text{w}_h^2.
\end{equation*}
In our design, the window sizes, the window sizes $\text{w}_h$ are typically kept small---for example, the last three scales use $\{3,5,7\}$---so $N_{\text{local}}$ is effectively a constant upper bound.
The computational complexity of $CSLA$ at scale $k$ therefore becomes
\begin{equation*}
    \mathcal{O}\left(dN_k(N_{\text{sink}}+N_{\text{local}})\right).
\end{equation*}
Moreover, since $N_{\text{sink}}$ covers only a few early scales (e.g., the \textbf{first 5 scales} contain just \textbf{121 KV tokens}), we have $N_{\text{sink}}+N_{\text{local}} \ll N_{\le k}$ at late high-resolution scales.
Therefore, compared with fully dense attention, $CSLA$ significantly reduces the computational cost.

\section{Implementation Details.}
The baseline VAR model, Infinity~\cite{infinity}, operates across a total of 13 resolution scales, corresponding to 13 iterations of next-scale autoregressive prediction.
We treat the \textbf{first 5 scales} as the \textbf{attention sink} and always preserve full attention computation over this region.
The sparsity configurations are set as follows:
(i) In the setting \textbf{without scale skipping}, we designate Scale 11 (with resolution $40\times40$) as the sparse decision scale, and apply hardware-efficient sparse computation to the last 2 scales for acceleration.
(ii) When \textbf{skipping the last two scales}~\cite{fastvar}, Scale 10 serves as the sparse decision scale, with sparse computation applied exclusively to Scale 11.
(iii) For the \textbf{sample-adaptive skipping strategy}~\cite{skipvar}, we similarly adopt Scale 10 as the sparse decision scale, sparsifying any subsequent scales that are not skipped.
Regarding the configuration of sparse attention modules, the application ratio of $CS^4A$ to $CSLA$ across all attention layers is approximately 6:4.
Unlike Infinity, HART~\cite{hart} operates over 14 scales. Following prior work~\cite{fastvar}, we do not apply scale skipping in any HART experiments, and set its sparse decision scale to Scale 11.
Finally, for the acceleration performance analysis, to ensure a fair comparison with the baseline model, we measure throughput using a batch size of 1.
%

\begin{figure*}[ht]
    \centering
    \includegraphics[width=\linewidth]{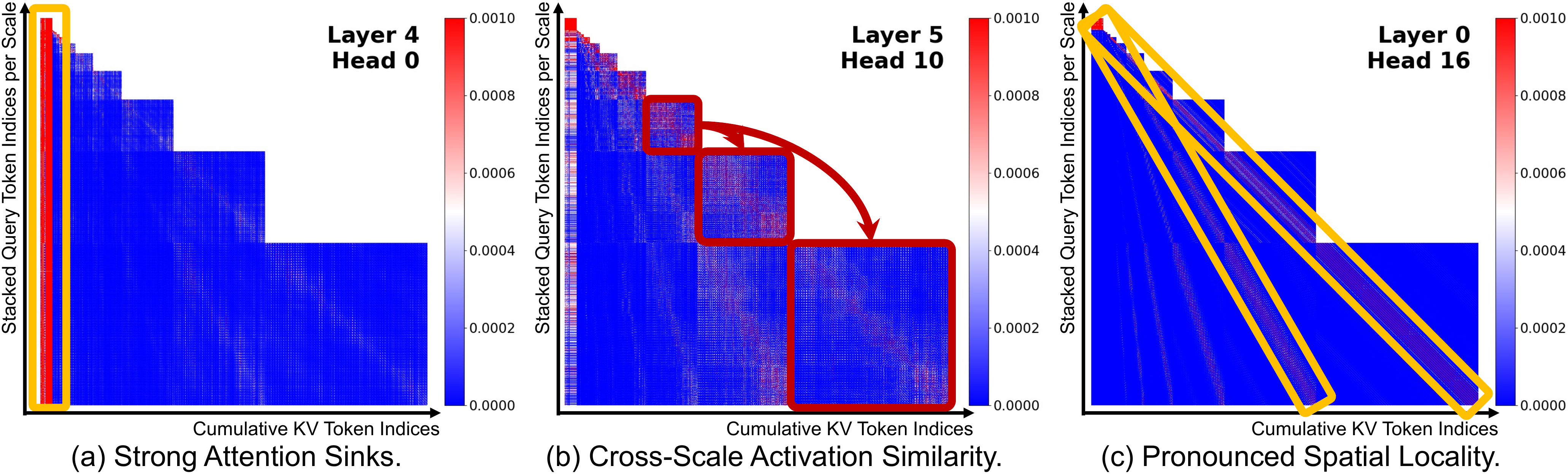}
    \caption{
        Visualization of attention activation patterns in the HART~\cite{hart} across different layers and heads.
    }
   \label{fig:hart_attn-map-ana}
\end{figure*}

\begin{table*}[ht]
    \centering
    \caption{
        Quantitative comparison on efficiency and quality on 1024$\times$1024 image generation.
        Note that the efficiency of HART baseline is tested under FlashAttention.
    }
    \label{tab:hart_comparison}
    \resizebox{\linewidth}{!}{%
        \begin{tabular}{lccccccccccccc}
            \toprule
            \multirow{2}{*}{\textbf{Methods}} & \multicolumn{4}{c}{\textbf{Acceleration}} & \multicolumn{2}{c}{\textbf{GPU Memory}} & \textbf{GenEval} & \textbf{DPG-Bench} & \textbf{HPSv2.1} & \textbf{ImageReward} & \multicolumn{3}{c}{\textbf{Quality Metrics}} \\
            \cmidrule(lr){2-5} \cmidrule(lr){6-7} \cmidrule(lr){8-8} \cmidrule(lr){9-9} \cmidrule(lr){10-10} \cmidrule(lr){11-11} \cmidrule(lr){12-14}
            & \#Scales $\downarrow$ & Speedup $\uparrow$ & Latency $\downarrow$ & Throughput $\uparrow$ & Torch Mem. $\downarrow$ & NV Mem. $\downarrow$ & Overall $\uparrow$ & Overall $\uparrow$ & Overall $\uparrow$ & Overall $\uparrow$ & PSNR $\uparrow$ & SSIM $\uparrow$ & LPIPS $\downarrow$ \\
            \midrule
            HART~\cite{hart} & 14 & 1.00$\times$ & 446.30ms & 2.24it/s & 19.35 & 27.93 & \textbf{0.509} & 74.754 & 29.070 & 0.661 & - & - & - \\
            FastVAR~\cite{fastvar} & 14 & 1.11$\times$ & 403.33ms & 2.48it/s & 19.33 & \textbf{25.72} & 0.504 & 74.762 & 27.680 & 0.602 & 20.850 & 0.704 & 0.316 \\
            \rowcolor{gray!15}
            \NAME (Ours) & 14 & \textbf{1.16$\times$} & \textbf{383.11ms} & \textbf{2.61it/s} & \textbf{19.29} & 26.98 & 0.507 & \textbf{75.625} & \textbf{29.140} & \textbf{0.680} & \textbf{21.157} & \textbf{0.735} & \textbf{0.240} \\
            \bottomrule
        \end{tabular}%
    }
\end{table*}

\section{Additional Experiments}

\paragraph{Comparison on DPG-Bench.}

To further validate the effectiveness of \NAME, we provide additional quantitative results on the DPG-Bench~\cite{dpg}, which evaluates prompt-following capabilities across multiple dimensions.
The results are summarized in Tab.~\ref{tab:dpg_results}.
For the 2B model, \NAME achieves a competitive overall score of \textbf{82.858} without skipping scales, surpassing the baseline and FastVAR while delivering a superior $\mathbf{1.38}\times$ speedup. 
When combined with scale-skipping strategies, our method maintains robust performance, achieving the highest score of \textbf{82.891} under the dynamic skipping setting with a $\mathbf{1.58\times}$ speedup, demonstrating an effective balance between acceleration and semantic fidelity.
For the larger 8B model, \NAME achieves an overall score of \textbf{86.411} without scale skipping, comparable to the baseline (86.440) and outperforming FastVAR and ScaleKV, all while providing a substantial $\mathbf{1.57\times}$ acceleration.
Notably, when integrated with scale-skipping strategies, \NAME consistently outperforms other baselines. 
In the ``skip last 2 scales'' setting, it attains the highest score of \textbf{86.468}---exceeding even the baseline---alongside a remarkable speedup.
The above results demonstrate the generality and effectiveness of our method.

\paragraph{Memory Consumption.}
We also provide a comprehensive memory comparison for generating 1024$\times$1024 images in the Tab.~\ref{tab:efficiency_memory}, including peak PyTorch active memory (Torch Mem.) and NVIDIA physical memory (NV Mem.).
The memory reduction in FastVAR and SkipVAR rely heavily on scale-skipping strategies.
Notably, FastVAR (w/o skip scales) exhibits memory usage nearly to the baseline, as the overhead of caching intermediate scale tokens for restoration offsets the benefits of token pruning.
While ScaleKV achieves the lowest memory, its design cannot be accelerated by modern efficiency attention kernels (e.g., FlashAttention), resulting in inference speeds significantly slower than the baseline.
Our sparse attention is optimized primarily for \textbf{inference latency} rather than memory minimization. Some reduction in physical memory originates from the efficient memory access strategies of the kernels.
As a `plug-and-play' module, \NAME is compatible with KV compression methods (e.g., retaining only sinks and local scales KV Cache, denoted as `+ KV Comp.' in Tab.~\ref{tab:efficiency_memory}).
We leave further dedicated memory optimizations to future work.

\paragraph{Generalizability of \NAME.}
We further validated \NAME on another representative VAR-based text-to-image model HART~\cite{hart}.
Our aim is to demonstrate that \NAME is not custom-designed for a specific VAR architecture (e.g., Infinity~\cite{infinity}), but rather serves as a plug-and-play module for accelerating visual generation models based on the next-scale prediction paradigm.
Fig.~\ref{fig:hart_attn-map-ana} illustrates the sparse activation patterns at each scale of the HART model, which exhibit three key properties similar to those observed in Infinity.
It is worth noting that since the first scale in HART is not a $1\times1$ token map but comprises text tokens with a fixed sequence length of 300, its attention sink phenomenon is even more stronger compared to Infinity.
Consequently, we adjust the configuration of the attention sink scales for our cross-scale sparse attention.
Specifically, in addition to the first 5 scales, the sink region must consistently include the initial text tokens.
%
Note that, following the setting of FastVAR, this model without scale skipping.
Shown in Tab.~\ref{tab:hart_comparison}, our \NAME achieves speedup and consistently outperforms FastVAR across 4 high-level benchmarks, even exceeding the baseline. 
Superior low-level metrics further confirm robust detail preservation.

\begin{figure*}[!ht]
    \centering
    \includegraphics[width=1\textwidth]{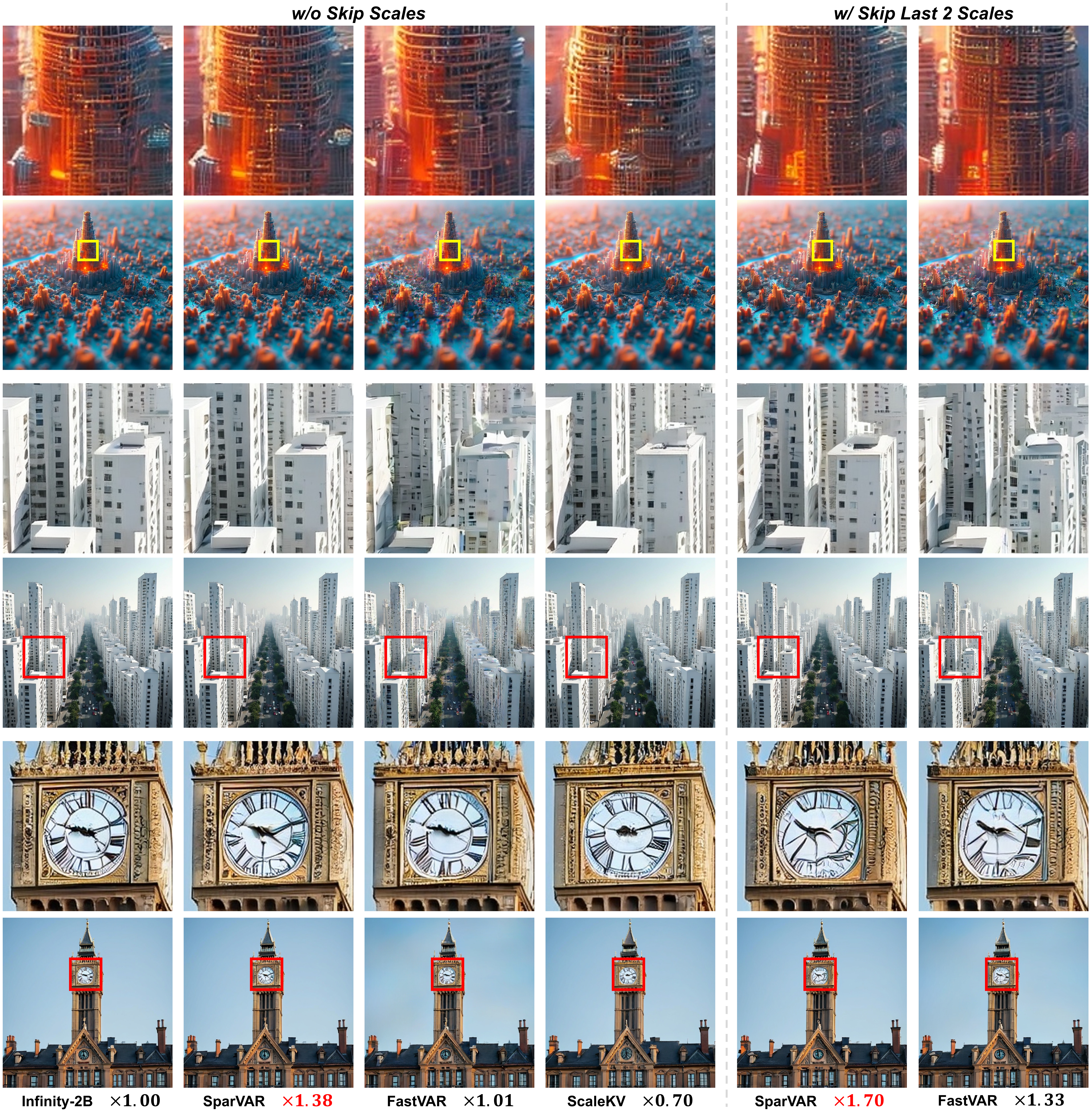}
    \caption{
        Qualitative comparison of complex scene generation on HPSv2.1~\cite{hpsv2} benchmark.
        Zoom in for fine-detail visualization.
    }
    \label{fig:display1}
\end{figure*}

\begin{figure*}[!ht]
    \centering
    \includegraphics[width=1\textwidth]{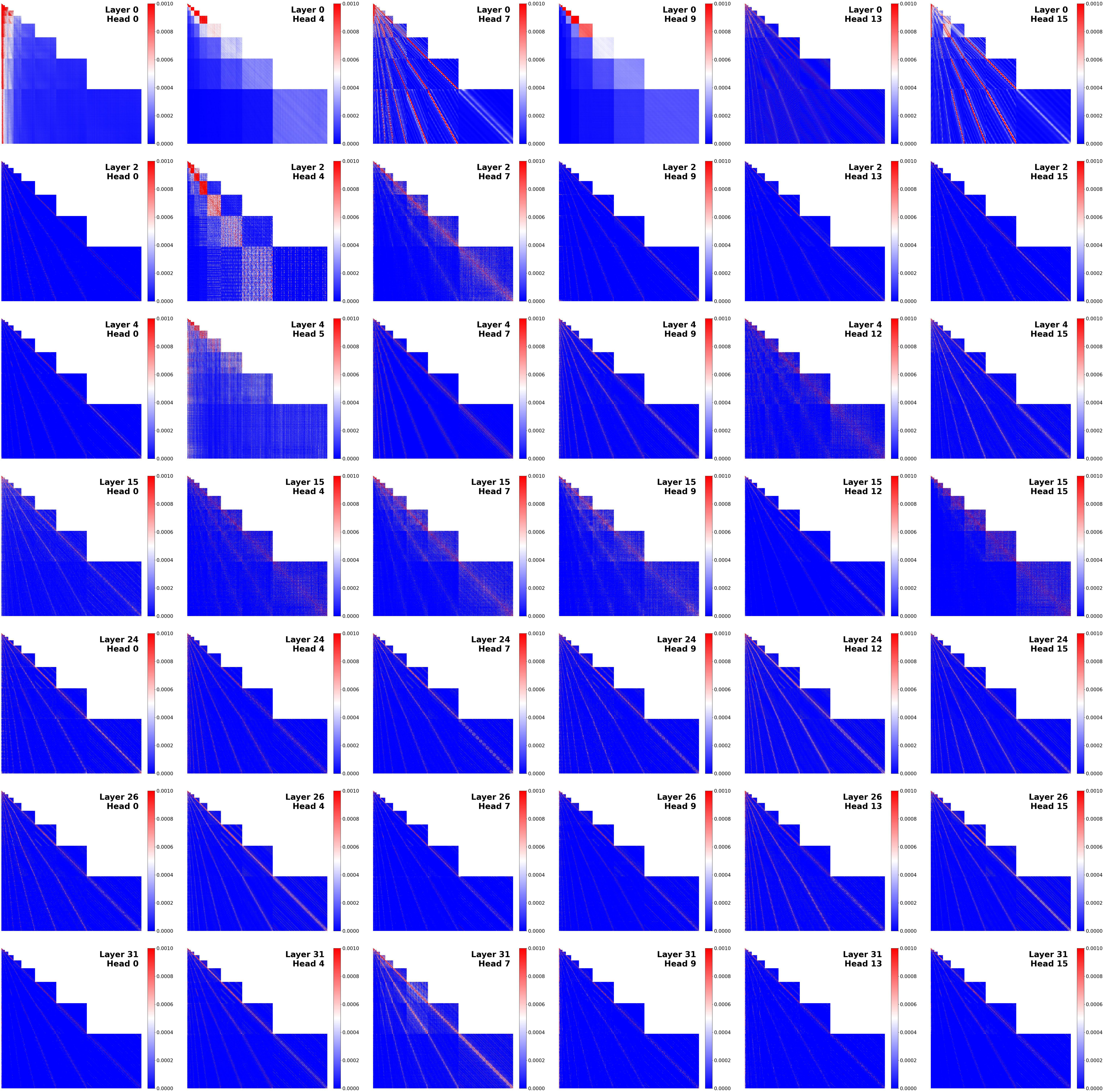}
    \caption{
        Visualization of the evolving attention activation patterns in VAR across layer depths.
    }
    \label{fig:depth}
\end{figure*}

\begin{figure}[!ht]
    \centering
    \includegraphics[width=1\linewidth]{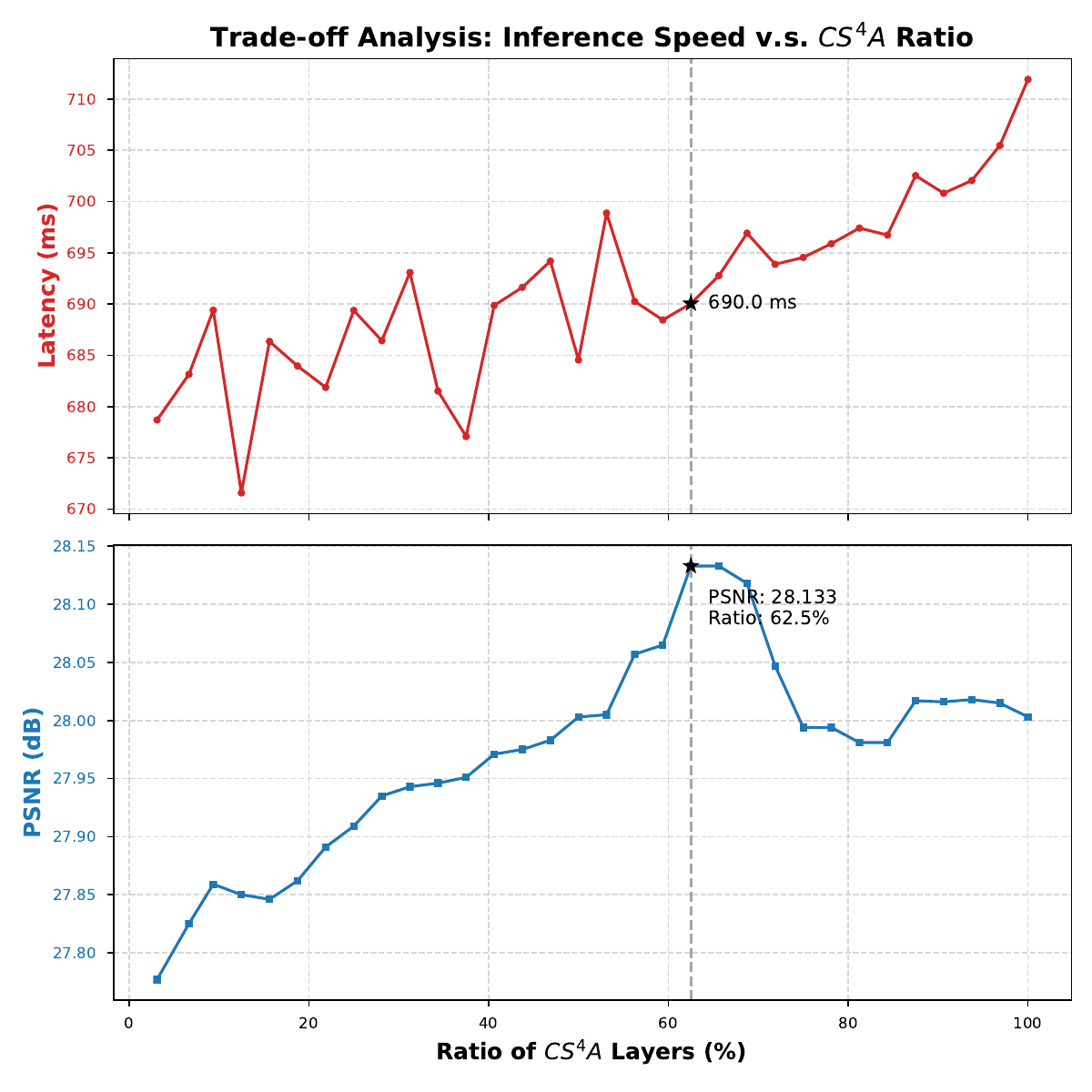}
    \caption{
        Configuration trade-off between $CS^4A$ and $CSLA$.
    }
    \label{fig:peibi}
\end{figure}

\paragraph{Qualitative Comparison.}

We compare \NAME against the baseline Infinity-2B and other training-free acceleration methods across complex generation scenarios from the HPSv2.1~\cite{hpsv2} benchmark.
As shown in Fig.~\ref{fig:display1}, in the setting without scale skipping, \NAME achieves a $1.38\times$ speedup while generating images with visual fidelity nearly identical to the baseline model. 
This is particularly evident in the second row, where our method accurately reconstructs the fine-grained square window panes of the white building---a structural detail that both FastVAR and ScaleKV fail to preserve, resulting in blurred or distorted features. 
Crucially, the robustness of \NAME extends to the aggressive skip-scale setting.
When skipping the last two high-resolution scales to achieve a $\mathbf{1.70\times}$ speedup, \NAME still successfully generates coherent structural details (e.g., the square windows), benefiting from the reinforced spatial locality in our sparse attention mechanism.
In contrast, FastVAR exhibits degradation, introducing severe artifacts and failing to maintain the geometric integrity of the scene.
These results qualitatively confirm that \NAME's cross-scale sparse attention is highly effective at compressing redundant computation while selectively preserving the critical high-frequency information necessary for photorealistic generation.
\textbf{Human evaluation of visualization} results is extremely important. 
We conducted a small-scale blind user study (constrained by budget) with 7 participants. The results indicate that $>70\%$ of evaluators preferred \NAME over other acceleration methods for its superior preservation of fine-grained details.

\paragraph{Optimal Configuration of Sparse Attention Modules.}
As illustrated in Fig.~\ref{fig:depth}, we further observe that in pretrained VAR models, the attention activation pattern evolves with layer depth, shifting from a relatively diverse distribution to a pronounced local pattern concentrated along the cross-scale diagonal.
Consequently, the dispersed activation patterns in shallower layers are better suited for $CS^4A$, which dynamically predicts sparse structures, whereas the fixed, strong locality in deeper layers is more effectively handled by the specialized $CSLA$.
To determine the optimal configuration, we conduct a systematic ablation study on the layer-wise allocation of $CS^4A$ and $CSLA$.
Starting from a baseline where all attention layers utilize $CSLA$, we progressively substitute them with $CS^4A$ from the shallowest to the deepest layers.

The quantitative results, illustrated in Fig.~\ref{fig:peibi}, reveal two key insights:
\textbf{1) Synergistic effect on generation quality.}
As the proportion of $CS^4A$ gradually increases, we observe a corresponding steady improvement in PSNR.
This trend aligns with our analysis, indicating that for the relatively dispersed sparse activation patterns in shallow layers, the cross-scale mapping of $CS^4A$ achieves superior pattern prediction compared to the fixed local window sparse of $CSLA$.
However, extending $CS^4A$ to the deepest layers (beyond 70\%) leads to a slight decline in PSNR. This suggests that deep layers are indeed specialized for high-frequency texture refinement, where the strong inductive bias of spatial locality enforced by $CSLA$ is more effective than the mapped sparse patterns.
\textbf{2) Latency-Quality trade-off.}
In terms of efficiency, replacing $CSLA$ with $CS^4A$ introduces a marginal increase in latency, primarily due to the overhead of index mapping and memory gathering compared to the highly optimized block-wise kernel of $CSLA$.
Nevertheless, the gain in visual fidelity (+0.35 dB PSNR) justifies this minimal cost.
Consequently, we adopt a hybrid configuration with about 60\% $CS^4A$ and 40\% $CSLA$ as the default setting for 2B and 8B models, ensuring optimal performance.



\end{document}